\documentclass[10pt,twocolumn,letterpaper]{article}

\usepackage{cvpr}              % To produce the CAMERA-READY version
\usepackage{graphicx}
\usepackage{amsmath}
\usepackage{amssymb}
\usepackage{booktabs}
\usepackage[linesnumbered,boxed,ruled,commentsnumbered]{algorithm2e}
\usepackage{multirow}
\usepackage{appendix}
\usepackage[accsupp]{axessibility}

\makeatletter
\@namedef{ver@everyshi.sty}{}
\makeatother
\usepackage{tikz}

\usepackage{silence}
\usepackage[pagebackref,breaklinks,colorlinks]{hyperref}

\usepackage[capitalize]{cleveref}
\crefname{section}{Sec.}{Secs.}
\Crefname{section}{Section}{Sections}
\Crefname{table}{Table}{Tables}
\crefname{table}{Tab.}{Tabs.}

\usepackage{lipsum}
\newcommand\blfootnote[1]{
    \begingroup
    \renewcommand\thefootnote{}\footnote{#1}
    \addtocounter{footnote}{-1}
    \endgroup
}

\def\cvprPaperID{1895} % *** Enter the CVPR Paper ID here
\def\confName{CVPR}
\def\confYear{2023}

\begin{document}

\twocolumn[{%
\renewcommand\twocolumn[1][]{#1}%
\title{ImageNet-E: Benchmarking Neural Network Robustness via Attribute Editing}
% \author{Xiaodan Li\\
% \and


\author{
Xiaodan Li\textsuperscript{1}, 
Yuefeng Chen\textsuperscript{1$\ast$},
Yao Zhu\textsuperscript{2}, 
Shuhui Wang\textsuperscript{3$\ast$}, 
Rong Zhang\textsuperscript{1},
Hui Xue\textsuperscript{1}\\
\textsuperscript{1}Alibaba Group \quad\quad \textsuperscript{2}Zhejiang University \quad\quad \textsuperscript{3}Inst. of Comput. Tech., CAS, China \\
\tt\small \{fiona.lxd, yuefeng.chenyf, stone.zhangr, hui.xueh\}@alibaba-inc.com\\
\tt\small ee$\_$zhuy@zju.edu.cn, wangshuhui@ict.ac.cn 
}

\maketitle

\vspace{-8mm}
\begin{center}
    \centering
    \captionsetup{type=figure}
    \includegraphics[width=0.9\textwidth]{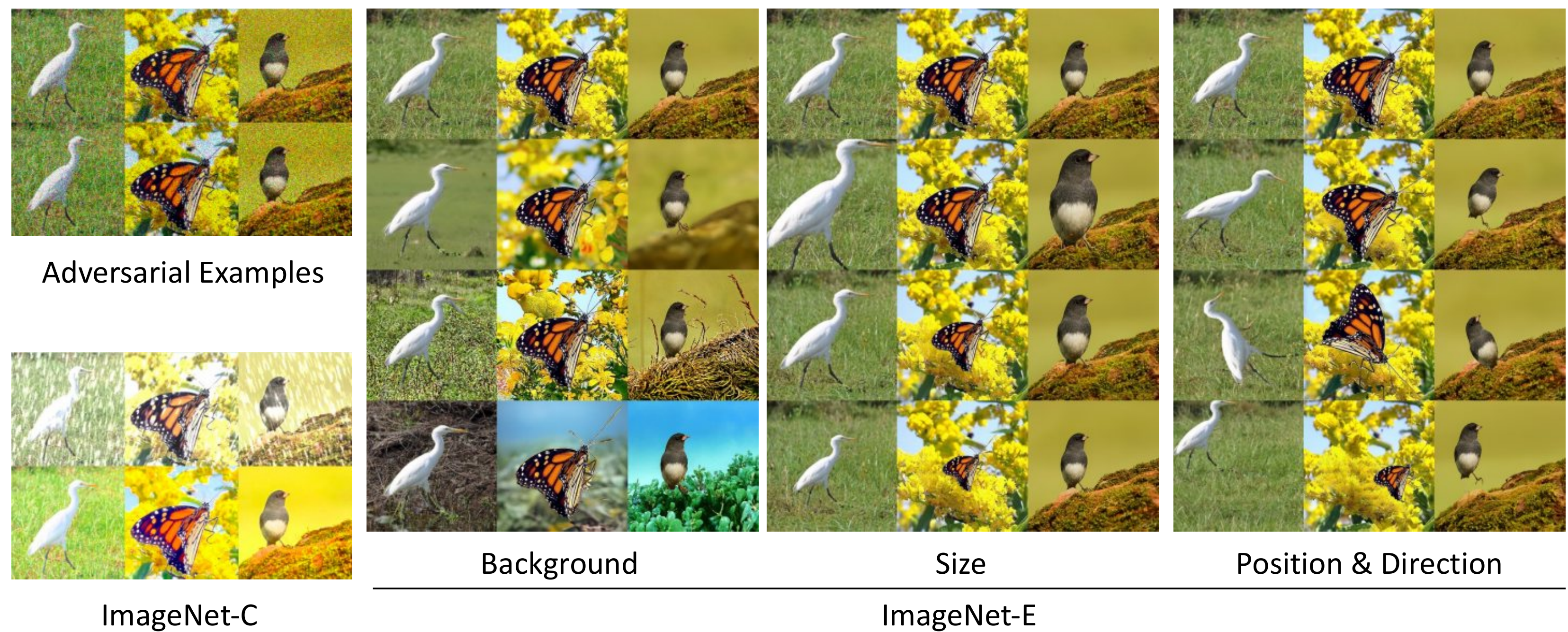}
    \vspace{-3mm}
    \captionof{figure}{Examples of the proposed ImageNet-E dataset. In contrast to adversarial examples or datasets like ImageNet-C~\cite{hendrycks2019benchmarking} who add perturbation or corruptions to original images, we edit the object attributes with controls of backgrounds, sizes, positions and directions.
    }
\label{fig:teaser}
\end{center}

}]
%   \includegraphics[width=\textwidth]{Figures/teaser.pdf}
%   \vspace{-7mm}
%   \caption{Examples of the proposed ImageNet-E dataset. In contrast to adversarial examples or datasets like ImageNet-C who add perturbation or corruptions to original images, we edit the object attributes with controls of backgrounds, sizes, positions and directions.}
%   \vspace{-2mm}
% \footnote{\textsuperscript{$\ast$}\  Corresponding author.}
% \affil[*]{Corresponding author}
\blfootnote{$\ast$  Corresponding author.}
\blfootnote{This research is supported in part by the National Key Research and Development Progrem of China under Grant No.2020AAA0140000.}

%%%%%%%%% ABSTRACT
\begin{abstract}
Recent studies have shown that higher accuracy on ImageNet usually leads to better robustness against different corruptions.
Therefore, in this paper, instead of following the traditional research paradigm that investigates new out-of-distribution corruptions or perturbations deep models may encounter, we conduct model debugging in in-distribution data to explore which object attributes a model may be sensitive to.
% To achieve this goal, we establish a rigorous benchmark named ImageNet-E(dit) for evaluating the image classifier robustness in terms of different object attributes.
% To support the data construction, we create a toolkit for object editing with controls of backgrounds, sizes, positions and directions.
To achieve this goal, we create a toolkit for object editing with controls of backgrounds, sizes, positions, and directions, and create a rigorous benchmark named ImageNet-E(diting) for evaluating the image classifier robustness in terms of object attributes.
With our ImageNet-E, we evaluate the performance of current deep learning models, including both convolutional neural networks and vision transformers.
% as well as the state-of-the-art self-supervised learning models. 
We find that most models are quite sensitive to attribute changes. A small change in the background can lead to an average of 9.23\% drop on top-1 accuracy. We also evaluate some robust models including both adversarially trained models and other robust trained models and find that some models show worse robustness against attribute changes than vanilla models.
Based on these findings, we discover ways to enhance attribute robustness with preprocessing, architecture designs, and training strategies.
We hope this work can provide some insights to the community and open up a new avenue for research in robust computer vision. 
The code and dataset are available at \small{\url{https://github.com/alibaba/easyrobust}}.
\end{abstract}

%%%%%%%%% BODY TEXT
\section{Introduction}
\label{sec:intro}

Deep learning has triggered the rise of artificial intelligence and has become the workhorse of machine intelligence. Deep models have been widely applied in various fields such as autonomous driving~\cite{huang2020survey}, medical science~\cite{litjens2017survey}, and finance~\cite{ozbayoglu2020deep}. With the spread of these techniques, the robustness and safety issues begin to be essential, especially after the finding that deep models can be easily fooled by negligible noises~\cite{goodfellow2014explaining}. As a result, more researchers contribute to building datasets for benchmarking model robustness to spot vulnerabilities in advance.

Most of the existing work builds datasets for evaluating the model robustness and generalization ability on out-of-distribution data~\cite{carlini2017adversarial,hendrycks2019benchmarking,kar20223d} using adversarial examples and common corruptions. 
For example, the ImageNet-C(orruption) dataset conducts visual corruptions such as Gaussian noise to input images to simulate the possible processors in real scenarios~\cite{hendrycks2019benchmarking}. ImageNet-R(enditions) contains various renditions~(\textit{e.g.}, paintings, embroidery) of ImageNet object classes~\cite{DeepAugment}.
As both studies have found that higher accuracy on ImageNet usually leads to better robustness against different domains ~\cite{hendrycks2019benchmarking,xiao2020noise}.
% We advocate that it is essential to conduct model debugging with the in-distribution dataset to provide clues for model accuracy improvement, besides exploring a new domain that models may confront. 
However, most previous studies try to achieve this in a top-down way, such as architecture design, exploring a better training strategy, \textit{etc}. We advocate that it is also essential to manage it in a bottom-up way, that is, conducting model debugging with the in-distribution dataset to provide clues for model repairing and accuracy improvement.
% For example, a bird in water is more likely to be mistaken as a fish due to the surroundings. That is the reason why more and more researches begin to study the causal and effect analysis~\cite{lv2022causality,cui2022stable}, which usually conduct experiments on domain generalization dataset. But how a deep model generalizes to different backgrounds is still unknown due to the lack of such an dataset.
For example, it is interesting to explore whether a bird with a water background can be recognized correctly even if most birds appear with trees or grasses in the training data.
Though this topic has been investigated in studies such as causal and effect analysis~\cite{cui2022stable}, the experiments and analysis are undertaken on domain generalization datasets. How a deep model generalizes to different backgrounds is still unknown due to the vacancy of a qualified benchmark.
Therefore, in this paper, we provide a detached object editing tool to conduct the model debugging from the perspective of object attribute and construct a dataset named ImageNet-E(diting).

The ImageNet-E dataset is a compact but challenging test set for object recognition that contains controllable object attributes including backgrounds, sizes, positions and directions, as shown in Fig.~\ref{fig:teaser}.
In contrast to ObjectNet~\cite{barbu2019objectnet} whose images are collected by their workers via posing objects according to specific instructions and differ from the target data distribution.
This makes it hard to tell whether the degradation comes from the changes of attribute or distribution.
Our ImageNet-E is automatically generated with our object attribute editing tool based on the original ImageNet.
% Write specific description of tool here.
% To be specific, we parse the image into class objects and backgrounds. 
Specifically, to change the object background, we provide an object background editing method that can make the background simpler or more complex based on diffusion models~\cite{sohl2015deep,ho2020denoising}.
In this way, one can easily evaluate how much the background complexity can influence the model performance. 
% We also edit object backgrounds in an adversarial way,\textit{i.e.}, making the input image away from its original class by changing the backgrounds.
To control the object size, position, and direction to simulate pictures taken from different distances and angles, 
% a controlling method is also provided.
an object editing method is also provided.
% Add the characteristic of ImageNet-E here.
With the editing toolkit, we apply it to the large-scale ImageNet dataset~\cite{russakovsky2015imagenet} to construct our ImageNet-E(diting) dataset. It can serve as a general dataset for benchmarking robustness evaluation on different object attributes.
% The proposed tool can also be adopted as a general dataset for validating the effectiveness of causal and effect analysis methods. To achieve this, we step further to generate a mini-dataset, which swaps the backgrounds of fishes with birds due to the reason that birds will not appear in water in most cases.

% Add experimental findings here.
With the ImageNet-E dataset, we evaluate the performance of current deep learning models, including both convolutional neural networks~(CNNs), vision transformers as well as the large-scale pretrained CLIP~\cite{CLIP}. We find that deep models are quite sensitive to object attributes. For example, when editing the background towards high complexity~(see Fig.~\ref{fig:teaser}, the $3$rd row in the background part), the drop in top-1 accuracy reaches 9.23\% on average.
We also find that though some robust models share similar top-1 accuracy on ImageNet, the robustness against different attributes may differ a lot.
% What we find very interesting and unexpected is that 
Meanwhile, some models, being robust under certain settings, 
% including adversarial trained models and non-adversarial trained models 
even show worse results than the vanilla ones on our dataset. This suggests that improving robustness is still a challenging problem and the object attributes should be taken into account.  
Afterward, we discover ways to enhance robustness against object attribute changes.
The main contributions are summarized as follows:
\vspace{-2mm}
\begin{itemize}
    \item We provide an object editing toolkit that can change the object attributes for manipulated image generation.
    \vspace{-5mm}
    \item We provide a new dataset called ImageNet-E that can be used for benchmarking robustness to different object attributes. It opens up new avenues for research in robust computer vision against object attributes.
    \vspace{-2mm}
    \item We conduct extensive experiments on ImageNet-E and find that models that have good robustness on adversarial examples and common corruptions may show poor performance on our dataset.
\end{itemize}

\begin{figure*}[t]
    \centering
    \includegraphics[width=\textwidth]{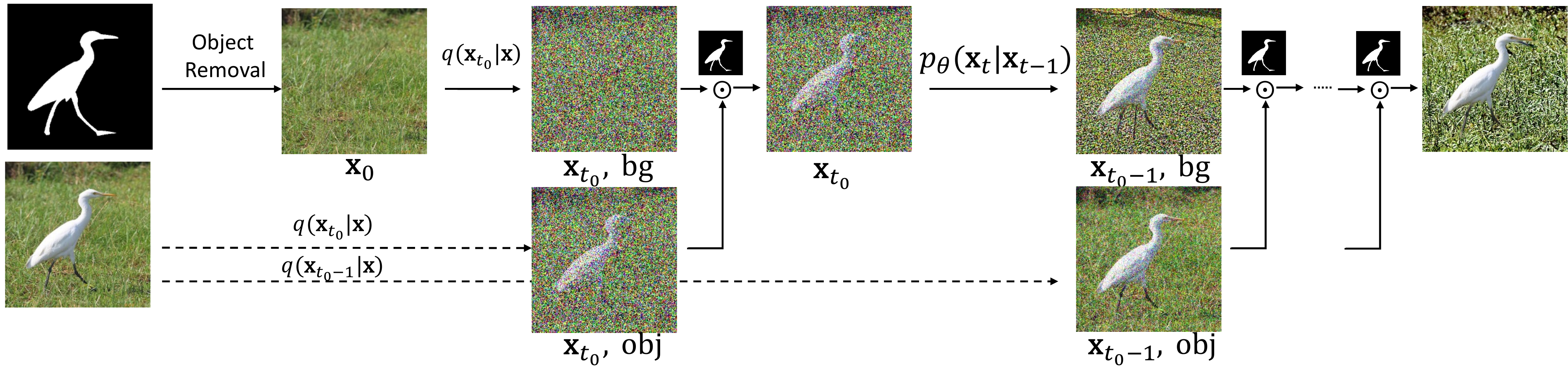}
\caption{Attribute editing with DDPMs. Give an input image and its corresponding object mask, the object is firstly removed with inpainting operation to get the pure background image. Then, we leverage the diffusion process to edit the background image $\mathbf{x}_0$ and object image coherently. $\odot$ denotes the element-wise blending of these two images using the object mask. For background editing, the background complexity objective function is added during the diffusion process (Alg.~\ref{algo:bg}, line $5$). For other object attributes editing, the object image needs to be transformed first (Alg.~\ref{algo_disjdecomp}, line $1$).}
    % \vspace{-5mm}
    \label{fig:DDPM_main_text}
\end{figure*}

% \begin{figure}[t]
%         % \vspace{-1cm}
% 		\centering
%         \includegraphics[height=4cm]{Figures/radar_3.pdf}
%         % \vspace{-0.25cm}
% 	    \caption{Top-1 accuracies on the original images~(Ori) and corresponding edited ones, including simple~(BG-S), complex~(BG-C) and adversarial~(BG-A) backgrounds, different object sizes including small~(S-S) and large~(S-L), and position~(P) editing.
% 	   % \vspace{-0.4cm}
% 	    }
% 		\label{fig:radar}
% \end{figure}

%------------------------------------------------------------------------

\section{Related Work}
\label{sec:related}

The literature related to attribute robustness benchmarks can be broadly grouped into the following themes: robustness benchmarks and attribute editing datasets. 
Existing robustness benchmarks such as ImageNet-C(orruption)~\cite{hendrycks2019benchmarking}, ImageNet-R(endition)~\cite{DeepAugment}, ImageNet-Stylized~\cite{SIN} and ImageNet-3DCC~\cite{kar20223d} mainly focus on the exploration of the corrupted or out-of-distribution data that models may encounter in reality. 
For instance, the ImageNet-R dataset contains various renditions (\textit{e.g.}, paintings, embroidery) of ImageNet object classes. ImageNet-C analyzes image models in terms of various simulated image corruptions~(\textit{e.g.}, noise, blur, weather, JPEG compression, \textit{etc.}).
Attribute editing dataset creation is a new topic and few studies have explored it before. 
% Among them, ObjectNet~\cite{barbu2019objectnet} and ImageNet-9~\cite{xiao2020noise} aim to facilitate the editing of object attribute. 
Among them, ObjectNet~\cite{barbu2019objectnet} and ImageNet-9~(\textit{a.k.a.} background challenge)~\cite{xiao2020noise} can be the representative. 
Specifically, ObjectNet collects a large real-world test set for object recognition with controls where object backgrounds, rotations, and imaging viewpoints are random. The images in ObjectNet are collected by their workers who image objects in their homes. It consists of $313$ classes which are mainly household objects. ImageNet-9 mainly creates a suit of datasets that help disentangle the impact of foreground and background signals on classification. To achieve this goal, it uses coarse-grained classes with corresponding rectangular bounding boxes to remove the foreground and then paste the cut area with other backgrounds. It can be observed that there lacks a dataset that can smoothly edit the object attribute.

\section{Preliminaries}

Since the editing tool is developed based on diffusion models,
let us first briefly review the theory of denoising diffusion probabilistic models~(DDPM)~\cite{sohl2015deep,ho2020denoising} and analyze how it can be used to generate images.
% Diffusion models are latent variable models of the form $p_\theta(\mathbf{x}_0) := \int p_\theta (\mathbf{x}_{0:T}) d \mathbf{x}_{1:T}$, where $x_1, ..., x_T$ are latents of the same dimensionality as the data $\mathbf{x}_0 \sim q(\mathbf{x}_0)$. The joint distribution $p_\theta (\mathbf{x}_{0:T})$ is called the reverse process, and it is defined as a Markov chain with learned Gaussian transitions starting at $p(\mathbf{x}_T) = \mathcal{N}(\mathbf{x}_T; 0,I)$:
% DDPMs learn to invert a parameterized markovian image noising process. Starting from isotropic gaussian noise samples, they transform them to samples from a training distribution, gradually removing the noise by an iterative diffusion process. DDPMs have recently been shown to generate high-quality images. 

According to the definition of the Markov Chain, one can always reach a desired stationary distribution from a given distribution along with the Markov Chain~\cite{geyer1992practical}. To get a generative model that can generate images from random Gaussian noises, one only needs to construct a Markov Chain whose stationary distribution is Gaussian distribution. 
This is the core idea of DDPM.
In DDPM, given a data distribution $\mathbf{x}_0 \sim q(\mathbf{x}_0)$, a forward noising process produces a series of latents $\mathbf{x}_1, ..., \mathbf{x}_T$ of the same dimensionality as the data $\mathbf{x}_0$ by adding Gaussian noise with variance $\beta_t \in (0,1)$ at time $t$:
\vspace{-2mm}
\begin{equation}
    q(\mathbf{x}_t | \mathbf{x}_{t-1}) = \mathcal{N}(\sqrt{1-\beta_t}\mathbf{x}_{t-1}, \beta_t \mathbf{I}), s.t.~ 0<\beta_t < 1,
\end{equation}
where $\beta_t$ is the diffusion rate.
% Since the mean and variance of this conditional Gaussian distribution are linear functions, it is easy to prove that 
Then the distribution $q(\mathbf{x}_t | \mathbf{x}_0)$ at any time $t$ is:
\begin{equation}
    q(\mathbf{x}_t | \mathbf{x}_0) = \mathcal{N}(\sqrt{\bar{\alpha}_t}, (1-\bar{\alpha}_t)\mathbf{I}),~ \mathbf{x}_t = \sqrt{\bar{\alpha}_t} \mathbf{x}_0 + {\sqrt{1-\bar{\alpha}_t}} \epsilon
    \label{eq:forward}
\end{equation}
where $\bar{\alpha}_t = \prod_{s=1}^t(1-\beta_t)$, $\epsilon \sim \mathcal{N}(0, \mathbf{I})$. It can be proved that $\lim_{t \to \infty}q(\mathbf{x}_t) = \mathcal{N}(0,\mathbf{I})$. In other words, we can map the original data distribution into a Gaussian distribution with enough iterations. Such a stochastic forward process is named as diffusion process since what the process $q(\mathbf{x}_t | \mathbf{x}_{t-1})$ does is adding noise to $\mathbf{x}_{t-1}$.

To draw a fresh sample from the distribution $q(\mathbf{x}_0)$, the Markov process is reversed. That is, beginning from a Gaussian noise sample $\mathbf{x}_T \sim \mathcal{N}(0, \mathbf{I})$, a reverse sequence is constructed by sampling the posteriors $q(\mathbf{x}_{t-1} | \mathbf{x}_t)$.
% , which were also found to be Gaussian distributions.
To approximate the unknown function $q(\mathbf{x}_{t-1} | \mathbf{x}_t)$, in DDPMs, a deep model $p_\theta$ is trained to predict the mean and the covariance of $\mathbf{x}_{t-1}$ given $\mathbf{x}_t$ instead. Then the $\mathbf{x}_{t-1}$ can be sampled from the normal distribution defined as:
\begin{equation}
    p_\theta(\mathbf{x}_{t-1}| \mathbf{x}_{t}) = \mathcal{N}(\mu_\theta(\mathbf{x}_t, t), \Sigma_\theta(\mathbf{x}_t, t)).
    \label{eq:denoise}
\end{equation}

In stead of inferring $\mu_\theta(\mathbf{x}_t, t)$ directly, \cite{ho2020denoising} propose to predict the noise $\epsilon_\theta(\mathbf{x}_t, t)$ which was added to $\mathbf{x}_0$ to get $\mathbf{x}_t$ with Eq.~\eqref{eq:forward}. Then $\mu_\theta(\mathbf{x}_t, t)$ is:
% may be derived using Bayes' theorem:
\begin{equation}
    \mu_\theta(\mathbf{x}_t, t) = \frac{1}{\sqrt{\bar{\alpha}_t}} \left(\mathbf{x}_t - \frac{\beta_t}{\sqrt{1-\bar{\alpha}_t}} \epsilon_\theta(\mathbf{x}_t, t)\right).
    \label{eq:mu}
\end{equation}
\cite{ho2020denoising} keep the value of $\Sigma_\theta(\mathbf{x}_t, t)$ to be constant. As a result, given a sample $\mathbf{x}_t$ at time $t$, with a trained model that can predict the noise $\epsilon_\theta(\mathbf{x}_t, t)$, we can get $\mu_\theta(\mathbf{x}_t, t)$ according to Eq.~\eqref{eq:mu} to reach the $\mathbf{x}_{t-1}$ with Equation~\eqref{eq:denoise} and eventually we can get to $\mathbf{x}_0$.
% , but it was later shown that it is better to learn it by a neural network that interpolates between the upper and lower bounds for the fixed covariance proposed by Ho \textit{et al}.

Previous studies have shown that diffusion models can achieve superior image generation quality compared to the current state-of-the-art generative models~\cite{avrahami2022blended}. Besides, there have been plenty of works on utilizing the DDPMs to generate samples with desired properties, such as semantic image translation~\cite{meng2021sdedit}, high fidelity data generation from low-density regions~\cite{sehwag2022generating}, \textit{etc}. 
In this paper, we also choose the DDPM adopted in \cite{avrahami2022blended} as our generator.
% due to its powerful generation ability.
% the properties that the generation process in DDPM is iterative and thus can be guided by a designed direction.

\section{Attribute Editing with Diffusion Models and ImageNet-E}

Most previous robustness-related work has focused on the important challenges of robustness on adversarial examples~\cite{carlini2017adversarial}, common corruptions~\cite{hendrycks2019benchmarking}. They have found that higher clean accuracy usually leads to better robustness. 
Therefore, instead of exploring a new corruption that models may encounter in reality, we pay attention to the model debugging in terms of object attributes, hoping to provide new insights to clean accuracy improvement.
% In contrast, we pay attention to the semantic robustness on objects with different backgrounds and sizes.
% We develop an object attribute editing tool for generating images while maintaining their semantic meaning. 
In the following, we describe our object attribute editing tool and the generated ImageNet-E dataset in detail. The whole pipeline can be found in Fig.~\ref{fig:DDPM_main_text}.

\begin{figure*}[t]
    \centering
    \includegraphics[height=4.5cm, width=\textwidth]{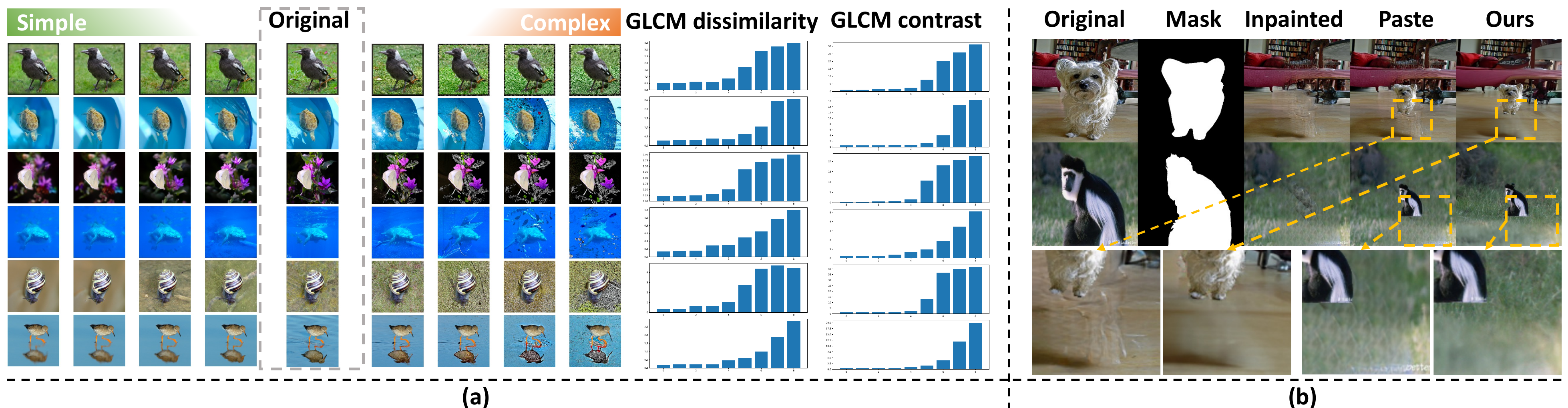}
    \vspace{-5mm}
    \caption{(a) Images generated with the proposed background complexity editing method. (b) Edited images with size changing. The Fréchet inception distance~(FID) for pasting is 50.64 while it is 32.59 for ours, indicating the effectiveness of the leveraging of DDPMs.}
    % \vspace{-5mm}
    \label{fig:bg_complex}
\end{figure*}

\subsection{Object Attribute Editing with Diffusion Models}

\noindent
\textbf{Background editing.} 
Most existing corruptions conduct manipulations on the whole image,
% for example, adversarial attacks are usually conducted on the whole image to get the adversarial examples. 
as shown in Fig.~\ref{fig:teaser}.
Compared to adding global corruptions that may hinder the visual quality, a more likely-to-happen way in reality is to manipulate the backgrounds to fool the model.
% However, the visual quality could be hindered by the global corruptions. 
% And we have no idea whether the performance decline is due to the quality degradation of the whole image or the impact of background change.
% In fact, a more likely-to-happen way in reality is to manipulate the backgrounds to attack the model.
Besides, it is shown that there exists a spurious correlation between labels and image backgrounds~\cite{geirhos2020shortcut}.
From this point, a background corruption benchmark is needed to evaluate the model's robustness.
However, the existing background challenge dataset achieves background editing with copy-paste operation, resulting an obvious artifacts in generated images~\cite{xiao2020noise}. This may leave some doubts about whether the evaluation is precise since the dataset's distribution may have changed.
To alleviate this concern, we adopt DDPM approach to incorporate background editing by adding a guiding loss that can lead to backgrounds with desired properties to make the generated images stay in/close to the original distribution.
Specifically, we choose to manipulate the background in terms of texture complexity due to the hypothesis that an object should be observed more easily from simple backgrounds than from complicated ones.
In general, the texture complexity can be evaluated with the gray-level co-occurrence matrix~(GLCM)~\cite{GLCM}, which calculates the gray-level histogram to show the texture characteristic.
However, the calculation of GLCM is non-differentiable, thus it cannot serve as the conditional guidance of image generation.
We hypothesize that a complex image should contain more frequency components in its spectrum and higher amplitude indicates greater complexity.
Thus, we define the objective of complexity as:
\vspace{-4mm}
\begin{equation}
    % \mathcal{L}_{c} = \sum abs(\mathcal{A}(\mathcal{F}(\mathbf{x}))),
    \mathcal{L}_{c} = \sum \left| \mathcal{A}(\mathcal{F}(\mathbf{x})) \right|,
\end{equation}
% \vspace{-2mm}
where $\mathcal{F}$ is the Fourier transform~\cite{bochner1949fourier}, $\mathcal{A}$ extracts the amplitude of the input spectrum. $\mathbf{x}$ is the evaluated image.
% $\mathbf{x}_0(\mathbf{x}_t, t))$ is the predicted $\mathbf{x}_0$ given $\mathbf{x}_t$ at time $t$.
Since minimizing this loss helps us generate an image with desired properties and should be conducted on the $\mathbf{x}_0$,
we need a way of estimating a clean image $\mathbf{x}_0$ from each noisy latent representation $\mathbf{x}_t$ during the denoising diffusion process. Recall that the process estimates at each step the noise $\epsilon_\theta(\mathbf{x}_t, t)$ added to $\mathbf{x}_0$ to obtain $\mathbf{x}_t$. Thus, 
% given an image $\mathbf{x}_t$ at time $t$, 
$\hat{\mathbf{x}}_0$ can be estimated via Equation~\eqref{eq:x_0}~\cite{avrahami2022blended}. The whole optimization procedure is shown in Algorithm~\ref{algo:bg}.
\begin{equation}
    \hat{\mathbf{x}}_0 = \frac{\mathbf{x}_t}{\sqrt{\bar{\alpha}_t}} - \frac{{\sqrt{1-\bar{\alpha}_t}} \epsilon_\theta(\mathbf{x}_t, t)}{\sqrt{\bar{\alpha}_t}}.
    \label{eq:x_0}
\end{equation}

As shown in Fig.~\ref{fig:bg_complex}(a), with the proposed method, when we guide the generation procedure with the proposed objective towards the complex direction, it will return images with visually complex backgrounds. 
% Each row shows the initial image $\textbf{x}_{t_0}$ at time $t_0$ (column 1) and corresponding generated images with different random noises. We also provide the GLCM dissimilarity of each image. 
% Each row shows the original image and its corresponding generated images under different weights. 
We also provide the GLCM dissimilarity and contrast of each image to make a quantitative analysis of the generated images.
A higher dissimilarity/contrast score indicates a more complex image background~\cite{GLCM}. 
It can be observed that the complexity is consistent with that calculated with GLCM, indicating the effectiveness of the proposed method.

% \begin{figure}[t]
%     \centering
%     \includegraphics[height=5cm]{Figures/size.pdf}
%     \caption{Edited images with size changing. The Fréchet inception distance~(FID) for pasting is 50.64 while it is 32.59 for ours, indicating the effectiveness of the leveraging of DDPMs.}
%     \label{fig:obj_size}
% \end{figure}

\noindent
\textbf{Controlling object size, position and direction.}
In general, the human vision system is robust to position, direction and small size changes. Whether the deep models are also robust to these object attribute changes is still unknown to researchers.
Therefore, we conduct the image editing with controls of object sizes, positions and directions to find the answer.
For a valid evaluation on different attributes, all other variables should remain unchanged, especially the background. 
Therefore, we first disentangle the object and background with the in-painting strategy provided by \cite{Zheng_2022_CVPR}.
Specifically, we mask the object area in input image $\mathbf{x}$. Then we conduct in-painting to remove the object and get the pure background image $\mathbf{x}^b$, as shown in Fig.~\ref{fig:bg_complex}(b) column $3$.
To realize the aforementioned object attribute controlling, we adopt the orthogonal transformation. 
% Denote $P$ as the pixel locations in image $\mathbf{x}$ where $P \in \mathbb{R}^{H \times W}$. $[H,W]$ is the size of $\mathbf{x}$. 
Denote $P$ as the pixel locations of object in image $\mathbf{x}$ where $P \in \mathbb{R}^{ 3 \times N_o}$. $N_o$ is the number of pixels belong to object and $p_i = [x_i, y_i, 1]^T$ is the position of object's $i$-th pixel.
$h' \in [0,H-h], w' \in [0,W-w]$ where $[x,y,w,h]$ stand for the enclosing rectangle of the object with mask $M$. 
Then the newly edited $\mathbf{x}[T_{\text{attribute}} \cdot P] = \mathbf{x}[P]$ and $M[T_{\text{attribute}} \cdot P] = M[P]$,
where 
\vspace{-2mm}
\begin{align}
\setlength{\arraycolsep}{1.2pt}
\resizebox{1.0\linewidth}{!}{
% \small
    $T_{\text{size}} = \left[
    \begin{array}{ccc}
    s & 0 & \Delta x \\
    0 & s & \Delta y \\
    0 & 0 & 1
    \end{array}
    \right], 
    T_{\text{position}} = \left[
    \begin{array}{ccc}
    1 & 0 & w' \\
    0 & 1 & h' \\
    0 & 0 & 1
    \end{array}
    \right], 
    T_{\text{direction}} = \left[
    \begin{array}{ccc}
    \cos{\theta} & \sin{\theta} & 0 \\
    -\sin{\theta} & \cos{\theta} & 0 \\
    0 & 0 & 1
    \end{array}
    \right].$}
\end{align}
% \vspace{-1mm}
where $s$ is the resize scale. $\theta$ is the rotation angle. $\Delta x = (1-s)\cdot(x+w/2), \Delta y = (1-s)\cdot(y+h/2)$.

With the background image $\mathbf{x}^b$ and edited object $\mathbf{x}^o$, a naive way is to place the object in the original image to the corresponding area of background image $\mathbf{x}^b$ as $M \odot \mathbf{x}^o + (1-M) \odot \mathbf{x}^b$.
% $M$ is the mask of the class object. 
However, the result generated in this manner may look disharmonic, lacking a delicate adjustment to blending them together. Besides, as shown in Fig.~\ref{fig:bg_complex}(b) column $3$, the object-removing operation may leave some artifacts behind, failing to produce a coherent and seamless result.
To deal with this problem, we leverage DDPM models to blend them at different noise levels along the diffusion process. Denote the image with desired object attribute as $\mathbf{x}^{o}$. Starting from the pure background image $\mathbf{x}^b$ at time $t_0$, at each stage, we perform a guided diffusion step with a latent $\mathbf{x}_t$ to obtain the $\mathbf{x}_{t-1}$ and at the same time, obtain a noised version of object image $\mathbf{x}_{t-1}^{o}$. Then the two latents are blended with the mask $M$ as $\mathbf{x}_{t-1} = M \odot \mathbf{x}_{t-1}^o + (1-M) \odot \mathbf{x}_{t-1}$.
The DDPM denoising procedure may change the background. Thus a proper initial timing is required to maintain a high resemblance to the original background. 
We set the iteration steps $t_0$ as 50 and 25 in Algorithm~\ref{algo:bg} and~\ref{algo_disjdecomp} respectively.

\begin{algorithm}[t]
% \begin{algorithmic} 
\small
\SetKwData{Left}{left}\SetKwData{This}{this}\SetKwData{Up}{up} \SetKwFunction{Union}{Union}\SetKwFunction{FindCompress}{FindCompress} \SetKwInOut{Input}{input}\SetKwInOut{Output}{output}
	
	\Input{source image $\mathbf{x}$, mask $M$, diffusion model $(\mu_\theta(\mathbf{x}_t), \Sigma_\theta(\mathbf{x}_t))$, $\bar{\alpha}_{t}$, $\lambda$, iteration steps $t_0$}
	
	\Output{edited image $\mathbf{x}_0$}
	 \BlankLine 
	 
% 	 \emph{special treatment of the first line}\; 
    %  $\mathbf{x}_0 \leftarrow \mathbf{x}$\;
     $\mathbf{x}_{t_0} \sim \mathcal{N}(\sqrt{\bar{\alpha}_{t_0}}\mathbf{x}, (1-\bar{\alpha}_{t_0})\mathbf{I})$\;
	 \For{$t\leftarrow t_0$ \KwTo $1$}{ 
% 	 	\emph{special treatment of the first element of line $i$}\; 
        % $\mu, \Sigma \leftarrow \mu_\theta(\mathbf{x}_t), \Sigma_\theta(\mathbf{x}_t)$\;
	 	$\hat{\mathbf{x}}_0 \leftarrow \frac{\mathbf{x}_t}{\sqrt{\bar{\alpha}_t}} - \frac{{\sqrt{1-\bar{\alpha}_t}} \epsilon_\theta(\mathbf{x}_t, t)}{\sqrt{\bar{\alpha}_t}}$\;
	 	$\nabla_{bg} \leftarrow \nabla_{\hat{\mathbf{x}}_0} \mathcal{L}_c (\hat{\mathbf{x}}_0)$\;
% 	 	$\mathbf{x}^b \sim \mathcal{N}(\mu - \lambda\Sigma\nabla_{bg}, \Sigma)$\;
	 	$\mathbf{x}^b_{t-1} \sim \mathcal{N}(\mu_\theta(\mathbf{x}_t) + \lambda\Sigma_\theta(\mathbf{x}_t)\nabla_{bg}, \Sigma_\theta(\mathbf{x}_t))$\;
	 	$\mathbf{x}^o \sim \mathcal{N}(\sqrt{\bar{\alpha}_t} \mathbf{x}, (1-\bar{\alpha}_t)\mathbf{I})$\;
	 	$\mathbf{x}_{t-1} \leftarrow M \odot \mathbf{x}^o + (1-M) \odot \mathbf{x}^b_{t-1}$\;
 	 } 
 	 	  \caption{Background editing}
 	 	  \label{algo:bg} 

% \end{algorithmic}

\end{algorithm}

\subsection{ImageNet-E dataset}
With the tool above, we conduct object attribute editing including background, size, direction and position changes based on the large-scale ImageNet dataset~\cite{russakovsky2015imagenet} and ImageNet-S~\cite{ImageNet-S}, which provides the mask annotation.
To guarantee the dataset quality, we choose the animal classes from ImageNet classes such as dogs, fishes and birds, since they appear more in nature without messy backgrounds. Classes such as stove and mortarboard are removed. 
% After this procedure, we get $373$ classes, resulting in $43520$ images in total. The image number of each class ranges from $3-16$. 
% Finally, our dataset consists of $47872$ images with $373$ classes.
Finally, our dataset consists of $47872$ images with $373$ classes based on the initial 4352 images, each of which is applied 11 transforms.
Detailed information can be found in Appendix~\ref{sec:ImageNet-E}. 
For background editing, we choose three levels of the complexity, including $\lambda=-20, \lambda=20$ and $\lambda=20\text{-adv}$ with adversarial guidance~(see Sec.\ref{appendix:bg} for details) instead of complexity. Larger $\lambda$ indicates stronger guidance towards high complexity. For the object size, we design four levels of sizes in terms of the object pixel rates~($=\text{sum}(M>0.5)/\text{sum}(M\geq0)$): $[\text{Full}, 0.1, 0.08, 0.05]$ where `Full' indicates making the object as large as possible while maintaining its whole body inside the image. Smaller rates indicate smaller objects. 
For object position, we find that some objects hold a high object pixel rate in the whole image, resulting in a small $H-h$. 
Take the first picture in Fig.~\ref{fig:bg_complex} for example, the dog is big and it will make little visual differences after position changing. 
Thus, we adopt the data whose pixel rate is 0.05 as the initial images for the position-changing operation.

% Our effort aims to give an editable image tool where the other attribute should stay unchanged when editing the target attribute to conduct the strict control variable methods for precise evaluation. 
% For example, the object background should remain the same when the object size changes when exploring how much the object sizes can influence on the model performance.
% hoping to provide some insights for clean accuracy improving. 

\begin{algorithm}[t]   
\small
\SetKwData{Left}{left}\SetKwData{This}{this}\SetKwData{Up}{up} \SetKwFunction{Union}{Union}\SetKwFunction{FindCompress}{FindCompress} \SetKwInOut{Input}{input}\SetKwInOut{Output}{output}
	
	\Input{source image $\mathbf{x}$, mask $M$, diffusion model $(\mu_\theta(\mathbf{x}_t), \Sigma_\theta(\mathbf{x}_t))$, $\bar{\alpha}_{t}$, iteration steps $t_0$, ratio $s$} 
	\Output{edited image $\mathbf{x}_0$}
	 \BlankLine 
	 \emph{$\mathbf{x}^b \leftarrow$ ObjectRemoving($\mathbf{x}, M$)};\\ %, $\mathbf{x}_0 \leftarrow \mathbf{x}^b$\;
	 \emph{$\mathbf{x}, M \leftarrow$  Rescale $(\mathbf{x}, M, s)$}\;
     $\mathbf{x}_{t_0} \sim \mathcal{N}(\sqrt{\bar{\alpha}_{t_0}}\mathbf{x}^b, (1-\bar{\alpha}_{t_0})\mathbf{I})$\;
	 \For{$t\leftarrow t_0$ \KwTo $1$}{ 
% 	 	\emph{special treatment of the first element of line $i$}\; 
        % $\mu, \Sigma \leftarrow \mu_\theta(\mathbf{x}_t), \Sigma_\theta(\mathbf{x}_t)$\;
% 	 	$\hat{\mathbf{x}}_0 \leftarrow \frac{\mathbf{x}_t}{\sqrt{\bar{\alpha}_t}} - \frac{{\sqrt{1-\bar{\alpha}_t}} \epsilon_\theta(\mathbf{x}_t, t)}{\sqrt{\bar{\alpha}_t}}$\;
% 	 	$\nabla_{bg} \leftarrow \nabla_{\hat{\mathbf{x}}_0} \mathcal{L}_c (\hat{\mathbf{x}}_0)$\;
% 	 	$\mathbf{x}^b \sim \mathcal{N}(\mu, \Sigma)$\;
	 	$\mathbf{x}^b_{t-1} \sim \mathcal{N}(\mu_\theta(\mathbf{x}_t), \Sigma_\theta(\mathbf{x}_t))$\;
	 	$\mathbf{x}^o \sim \mathcal{N}(\sqrt{\bar{\alpha}_t} \mathbf{x}, (1-\bar{\alpha}_t)\mathbf{I})$\;
	 	$\mathbf{x}_{t-1} \leftarrow M \odot \mathbf{x}^o + (1-M) \odot \mathbf{x}^b_{t-1}$\;
     \vspace{0.135cm}
 	 } 
 	 	  \caption{Object size controlling}
 	 	  \label{algo_disjdecomp} 
\end{algorithm}

\begin{figure*}[t]
    \centering
    \includegraphics[height=4.5cm]{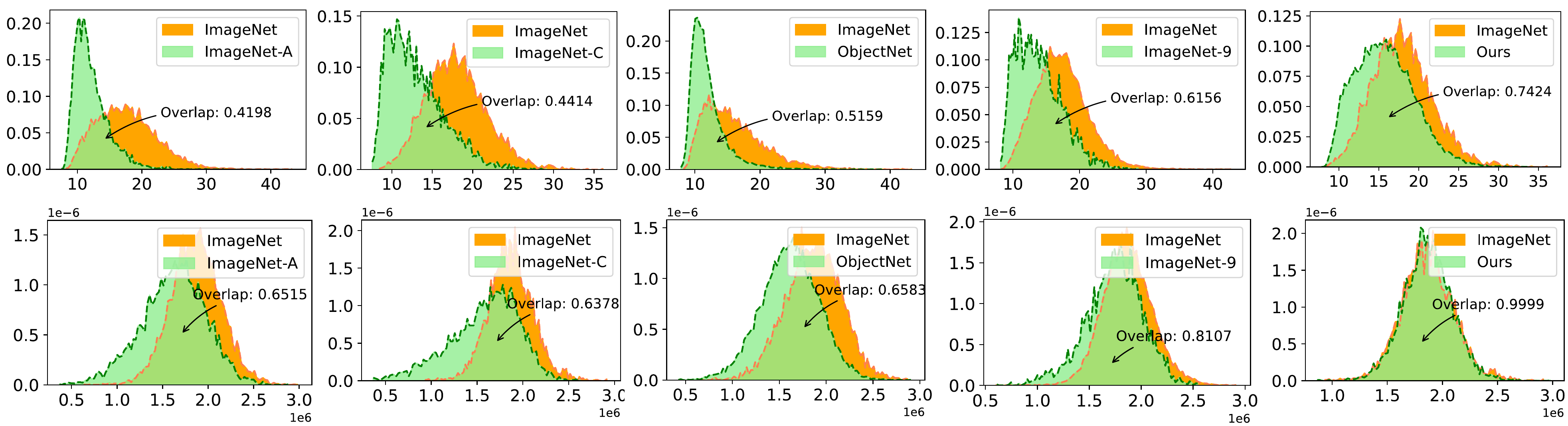}
    \caption{Distributions of ID score of different datasets in terms of the quantities in Energy (the first row) and GradNorm (the second row) for in-distribution~(ImageNet) and other datasets. Higher overlap indicates greater proximity to ImageNet.}
    \vspace{-3mm}
    \label{fig:OOD}
\end{figure*}

In contrast to benchmarks like ImageNet-C~\cite{hendrycks2019benchmarking} giving images from different domains so that the model robustness in these situations may be assessed, our effort aims to give an editable image tool that can conduct model debugging with in-distribution~(ID) data, in order to identify specific shortcomings of different models and provide some insights for clean accuracy improving.
Thus, the data distribution should not differ much from the original ImageNet.
We choose the out-of-distribution~(OOD) detection methods Energy~\cite{liu2020energy} and GradNorm~\cite{huang2021importance} to evaluate whether our editing tool will move the edited image out of its original distribution. 
These OOD detection methods aim to distinguish the OOD examples from the ID examples. The results are shown in Fig.~\ref{fig:OOD}. $x$-axis is the ID score in terms of the quantities in Energy and GradNorm and $y$-axis is the frequency of each ID score. A high ID score indicates the detection method takes the input sample as the ID data.
Compared to other datasets, our method barely changes the data distribution under both Energy~(the 1st row) and GradNorm~(the 2nd row) evaluation methods. 
Besides, the Fréchet inception distance~(FID)~\cite{FID} for our ImageNet-E is 15.57 under the random background setting, while it is 34.99 for ImageNet-9 (background challenge).
These all imply that our editing tool can ensure the proximity to the original ImageNet, thus can give a controlled evaluation on object attribute changes.
To find out whether the DDPM will induce some degradation to our evaluation, we have conducted experiment in Tab.~\ref{tab:arch} with the setting $\lambda=0$ during background editing. This operation will first add noises to the original and then denoise them.
It can be found in ``Inver'' column that the degradation is negligible compared to degradation induced by attribute changes.

\section{Experiments}
\label{Experiment}

% \subsection{Experimental settings}
We conduct evaluation experiments on various architectures including both CNNs~(ResNet~(RN)~\cite{resnet}, DenseNet~\cite{huang2017densely}, EfficientNet~(EF)~\cite{tan2019efficientnet}, ResNest~\cite{zhang2022resnest}, ConvNeXt~\cite{liu2022convnet}) and transformer-based models~(Vision-Transformer~(ViT)~\cite{vit}, Swin-Transformer~(Swin)~\cite{liu2021swin}). Other state-of-the-art models that trained with extra data such as CLIP~\cite{CLIP}, EfficientNet-L2-Noisy-Student~\cite{NT} are also evaluated in the Appendix. Apart from different sizes of these models, we have also evaluated their adversarially trained versions for comprehensive studies.
We report the drop of top-1 accuracy as metric based on the idea that the attribute changes should induce little influence to a robust trained model. More experimental details and results of top-1 accuracy can be found in the Appendix.
% Besides, non-adversarially robust models that show good performance on other robustness benchmarks are also evaluated, including SIN~\cite{SIN}, BebiasedCNN~\cite{debiased}, Augmix~\cite{hendrycks2020Augmix}, ANT~\cite{ANT}, DeepAugment~\cite{DeepAugment}.
% Our code is implemented based on PyTorch~\cite{paszke2019pytorch}.

% \subsection{Proximity Evaluation} % sample quality
% \textbf{Sample quality and proximity.}

\subsection{Robustness evaluation}

\noindent
\textbf{Normally trained models.}
To find out whether the widely used models in computer vision have gained robustness against changes on different object attributes, we conduct extensive experiments on different models.
As shown in Tab.~\ref{tab:arch}, when only the background is edited towards high complexity, the average drop in top-1 accuracy is 9.23\%~($\lambda=20$). 
% Note that this drop rate is not low since our generated images are close to in-distribution data.
% However, for human beings, the change is imperceptible, as shown in Fig.~\ref{fig:teaser}, row $3$. 
This indicates that most models are sensitive to object background changes.
Other attribute changes such as size and position can also lead to model performance degradation. For example, when changing the object pixel rate to $0.05$, as shown in Fig.~\ref{fig:teaser} row $4$ in the `size' column, while we can still recognize the image correctly, the performance drop is 18.34\% on average.
We also find that the robustness under different object attributes is improved along with improvements in terms of clean accuracy~(Original) on different models. Accordingly, a switch from an RN50 (92.69\% top-1 accuracy) to a Swin-S (96.21\%) leads to the drop in accuracy decrease from 15.72\% to 10.20\% on average. 
By this measure, models have become more and more capable of generalizing to different backgrounds, which implies that they indeed learn some robust features. This shows that object attribute robustness can be a good way to measure future progress in representation learning.
% An interesting finding is that the DenseNet shows higher robustness against background changes even if its accuracy on original images is lower than EfficientNet.
We also observe that larger networks possess better robustness on the attribute editing. For example, swapping a Swin-S (96.21\% top-1 accuracy) with the larger Swin-B (95.96\% top-1 accuracy) leads to the decrease of the dropped accuracy from 10.20\% to 8.99\% when $\lambda=20$. In a similar fashion, a ConvNeXt-T (9.32\% drop) is less robust than the giant ConvNeXt-B (7.26\%). Consequently, models with even more depth, width, and feature aggregation may attain further attribute robustness. 
% We also test two well-known self-supervised learning methods MAE~\cite{mae} and MoCo-v3~\cite{MoCo}. MAE is a famous self-supervised learning method that learns to reconstruct the input image with few patches while MoCo-v3 is a representative work for contrastive learning methods. We find that the MAE method is quite robust in our ImageNet-E dataset while MoCo-v3 shows poor performance. We think this 
Previous studies~\cite{kumar2022fine} have shown that zero-shot CLIP exhibits better out-of-distribution robustness than the finetuned CLIP, which is opposite to our ImageNet-E as shown in Tab.~\ref{tab:arch}. This may serve as the evidence that our ImageNet-E has a good proximity to ImageNet. We also find that compared with fully-supervised trained model under the same backbone~(ViT-B), the CLIP fails to show a better attribute robustness. We think this may be caused by that the CLIP has spared some capacity for OOD robustness.

\begin{table*}[ht]
\caption{Evaluations with different state-of-the-art models in terms of Top-1 accuracy and the corresponding drop of accuracy under background changes, size changes, random position (rp) and random direction (rd).}
\resizebox{1.0\linewidth}{!}{
\begin{tabular}{c|c|ccccc|cccc|c|c|c}
\toprule
% \multirow{2}{*}{Models} & \multicolumn{1}{c|}{\multirow{2}{*}{Original}} & \multicolumn{1}{c|}{\multirow{2}{*}{Inver}} & \multicolumn{4}{c|}{Background changes}                                                                                                   & \multicolumn{4}{c|}{Size changes}                                                                         & \multicolumn{1}{c|}{Position} & \multicolumn{1}{c|}{Direction} & \multirow{2}{*}{Avg.} \\ \cline{4-13}
\multirow{2}{*}{Models} & \multicolumn{1}{c|}{\multirow{2}{*}{Original}} & \multicolumn{5}{c|}{Background changes}                                                                                                   & \multicolumn{4}{c|}{Size changes}                                                                         & \multicolumn{1}{c|}{Position} & \multicolumn{1}{c|}{Direction} & \multirow{2}{*}{Avg.} \\ \cline{3-13}
                        & \multicolumn{1}{c|}{}                          & \multicolumn{1}{c}{Inver}                       & \multicolumn{1}{c}{$\lambda=-20$} & \multicolumn{1}{c}{$\lambda=20$} & \multicolumn{1}{c}{$\lambda=20$-adv} & \multicolumn{1}{c|}{Random} & \multicolumn{1}{c}{Full} & \multicolumn{1}{c}{0.1} & \multicolumn{1}{c}{0.08} & \multicolumn{1}{c|}{0.05} & \multicolumn{1}{c|}{rp}       & \multicolumn{1}{c|}{rd}        &                       \\ \hline\hline
RN50                    & 92.69\%                                       & 1.97\%                                     & 7.30\%                            & 13.35\%                          & 29.92\%                              & 13.34\%                    & 2.71\%                   & 7.25\%                  & 10.51\%                  & 21.26\%                  & 26.46\%                      & 25.12\%                       & 15.72\%               \\
DenseNet121             & 92.10\%                                       & 1.49\%                                     & 6.29\%                            & 9.00\%                           & 29.20\%                              & 12.43\%                    & 3.50\%                   & 7.00\%                  & 10.68\%                  & 21.55\%                  & 26.53\%                      & 23.64\%                       & 14.98\%               \\
EF-B0                   & 92.85\%                                       & 1.07\%                                     & 7.10\%                            & 10.71\%                          & 34.88\%                              & 15.64\%                    & 3.03\%                   & 8.00\%                  & 11.57\%                  & 23.28\%                  & 27.91\%                      & 19.11\%                       & 16.12\%               \\
ResNest50               & 95.38\%                                       & 1.44\%                                     & 6.33\%                            & 8.98\%                           & 26.62\%                              & 11.28\%                    & 2.53\%                   & 5.27\%                  & 8.01\%                   & 18.03\%                  & 21.37\%                      & 17.32\%                       & 12.57\%               \\
ViT-S                  & 94.14\%                                       & \textbf{0.82\%}                            & 6.42\%                            & 8.98\%                           & 31.12\%                              & 13.06\%                    & \textbf{0.80\%}          & 5.37\%                  & 8.59\%                   & 17.37\%                  & 22.86\%                      & 17.13\%                       & 13.17\%               \\
% ViT-S                   & 94.74\%                                       & 1.66\%                                     & 7.32\%                            & 10.64\%                          & 32.17\%                              & 14.39\%                    & 1.22\%                   & 7.10\%                  & 10.64\%                  & 20.29\%                  & 25.08\%                      & 17.22\%                       & 14.61\%               \\
Swin-S                  & \textbf{96.21\%}                              & 1.13\%                                     & 5.18\%                            & 7.33\%                           & 23.50\%                              & 9.31\%                     & 1.27\%                   & 4.21\%                  & 6.29\%                   & 14.16\%                  & 17.35\%                      & \textbf{13.42\%}              & 10.20\%               \\
ConvNeXt-T              & 96.07\%                                       & 1.43\%                                     & \textbf{4.69\%}                   & \textbf{6.26\%}                  & \textbf{19.83\%}                     & \textbf{7.93\%}            & 1.75\%                   & \textbf{3.28\%}         & \textbf{5.18\%}          & \textbf{12.76\%}         & \textbf{15.71\%}             & 15.78\%                       & \textbf{9.32\%}       \\
\hline\hline
RN101                   & 94.00\%                                       & 2.11\%                                     & 7.05\%                            & 11.62\%                          & 29.47\%                              & 13.57\%                    & 2.57\%                   & 6.81\%                  & 10.12\%                  & 20.65\%                  & 25.85\%                      & 24.42\%                       & 15.21\%               \\
DenseNet169             & 92.37\%                                       & 1.12\%                                     & 5.81\%                            & 8.43\%                           & 27.51\%                              & 11.61\%                    & 2.25\%                   & 6.90\%                  & 10.41\%                  & 20.59\%                  & 24.93\%                      & 20.68\%                       & 13.91\%               \\
EF-B3                   & 94.97\%                                       & 1.87\%                                     & 7.77\%                            & 8.40\%                           & 29.90\%                              & 12.92\%                    & 1.36\%                   & 6.80\%                  & 10.16\%                  & 21.36\%                  & 24.98\%                      & 17.24\%                       & 14.09\%               \\
ResNest101              & 95.54\%                                       & 1.10\%                                     & 5.58\%                            & 6.65\%                           & 23.03\%                              & 10.40\%                    & 1.35\%                   & 3.97\%                  & 6.53\%                   & 15.44\%                  & 19.11\%                      & 14.31\%                       & 10.64\%               \\
ViT-B                  & 95.38\%                                       & 0.83\%                                     & 5.32\%                            & 8.43\%                           & 26.60\%                              & 10.98\%                    & \textbf{0.62\%}                   & 4.00\%                  & 6.30\%                   & 14.51\%                  & 18.82\%                      & 14.95\%                       & 11.05\%               \\
% ViT-B                   & 95.66\%                                       & \textbf{0.68\%}                            & 5.21\%                            & 8.01\%                           & 24.25\%                              & 10.42\%                    & \textbf{0.46\%}          & 4.69\%                  & 6.57\%                   & 15.68\%                  & 19.47\%                      & \textbf{11.44\%}              & 10.62\%               \\
Swin-B                  & 95.96\%                                       & 0.79\%                                     & 4.46\%                            & 6.23\%                           & 21.44\%                              & 8.25\%                     & 0.99\%                   & 3.16\%                  & 5.04\%                   & 12.34\%                  & 15.38\%                      & \textbf{12.60\%}                       & 8.99\%                \\
ConvNeXt-B              & \textbf{96.42\%}                              & \textbf{0.69\%}                                     & \textbf{3.75\%}                   & \textbf{4.86\%}                  & \textbf{16.49\%}                     & \textbf{6.04\%}            & 0.99\%                   & \textbf{2.25\%}         & \textbf{3.36\%}          & \textbf{9.47\%}          & \textbf{12.40\%}             & 13.01\%                       & \textbf{7.26\%}       \\ \hline
CLIP-zeroshot           & 80.01\%                                       & 4.88\%                                     & 11.56\%                           & 15.28\%                          & 36.14\%                              & 20.09\%                    & 3.33\%                   & 12.67\%                 & 15.77\%                  & 25.31\%                  & 28.87\%                      & 21.57\%                       & 19.06\%               \\
CLIP-finetuned           & 93.68\%                                       & 2.17\%                                     & 9.82\%                            & 11.83\%                          & 38.33\%                              & 18.19\%                    & 9.06\%                   & 9.25\%                  & 12.67\%                  & 23.32\%                  & 28.56\%                      & 22.00\%                       & 18.30\%            
\\ \toprule

\end{tabular}}
\label{tab:arch}
\vspace{-3mm}
\end{table*}

\begin{figure}[!htb]
    \centering
    \includegraphics[height=3cm]{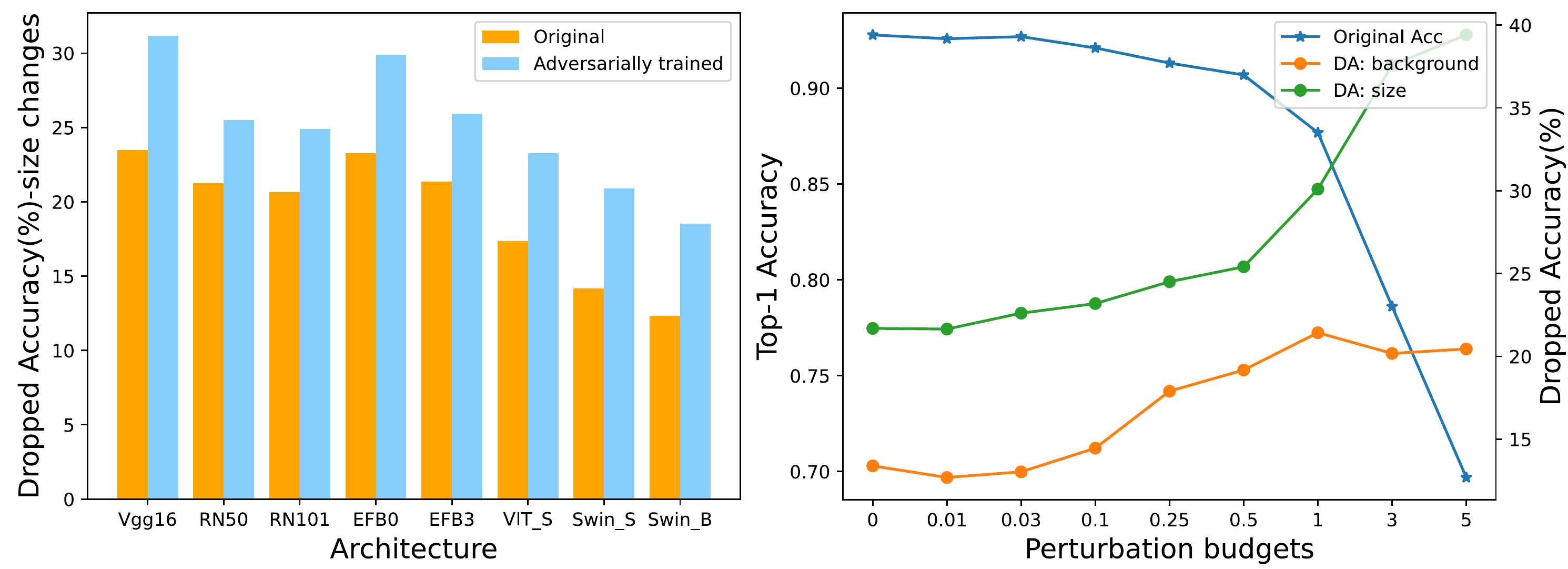}
\caption{Comparisons between vanilla models and adversarially trained models across different architectures in terms of size changes~(left). Evaluation of adversarial models (RN50) trained with different perturbation budgets is provided in the right figure.}
    \vspace{-3mm}
    \label{fig:adv}
\end{figure}

\noindent
\textbf{Adversarially trained models.}
% Adversarial training~\cite{salman2020adversarially} is the state-of-the-art method for improving the adversarial robustness of deep neural networks and has been widely studied.
Adversarial training~\cite{salman2020adversarially} is one of the state-of-the-art methods for improving the adversarial robustness of deep models and has been widely studied~\cite{bai2021recent}.
To find out whether they can boost the attribute robustness, we conduct extensive experiments in terms of different architectures and perturbation budgets~(constraints of $l_2$ norm bound).
As shown in Fig.~\ref{fig:adv}, 
% it is surprising to find that
the adversarially trained ones are not robust against attribute changes including both backgrounds and size-changing situations. The dropped accuracies are much greater compared to normally trained models.
As the perturbation budget grows, the situation gets worse. This indicates that adversarial training can do harm to robustness against attributes.

\subsection{Robustness enhancements}
Based on the above evaluations, we step further to discover ways to enhance the attribute robustness in terms of preprocessing, network design and training strategies. More details including training setting and numerical experimental results can be found in Appendix~\ref{sec:robust_enhance}.

\noindent
\textbf{Preprocessing.}
Given that an object can be inconspicuous due to its small size or subtle position, viewing an object at several different locations may lead to a more stable prediction. Having this intuition in mind, we perform the classical Ten-Crop strategy to find out if this operation can help to get a robustness boost. The Ten-Crop operation is executed by cropping all four corners and the center of the input image. We average the predictions of these crops together with their horizontal mirrors as the final result. We find this operation can contribute a 0.69\% and 1.24\% performance boost on top-1 accuracy in both background and size changes scenarios on average respectively.

\noindent
\textbf{Network designs.}
Intuitively, a robust model should tend to focus more on the object of interest instead of the background. Therefore, recent models begin to enhance the model by employing attention modules. Of these, the ResNest~\cite{zhang2022resnest} can be a representative. The ResNest is a modularized architecture, which applies channel-wise attention on different network branches to leverage their success in capturing cross-feature interactions and learning diverse representations. As it has achieved a great boost in the ImageNet dataset, it also shows superiority on ImageNet-E compared to ResNet. For example, a switch from RN50 decreases the average dropped accuracy from 15.72\% to 12.57\%. This indicates that the channel-wise attention module can be a good choice to improve the attribute robustness. 
Another representative model can be the vision transformer, which consists of multiple self-attention modules. To study whether incorporating transformer's self-attention-like architecture into the model design can help attribute robustness generalization, we establish a hybrid architecture by directly feeding the output of res\_3 block in RN50 into ViT-S as the input feature like ~\cite{bai2021transformers}. The dropped accuracy decreases by 1.04\% compared to the original RN50, indicating the effectiveness of the self-attention-like architectures.

\begin{table*}[t]
\caption{Evaluations with different robust models in terms of Top-1 accuracy and the corresponding dropped accuracy.}
\resizebox{1.0\linewidth}{!}{
\begin{tabular}{c|c|ccccc|cccc|c|c|c}
\toprule
\multirow{2}{*}{Architectures} & \multirow{2}{*}{Ori} &  \multicolumn{5}{c|}{Background changes}                                                   & \multicolumn{4}{c|}{Size changes}                                  & Position         & Direction & \multirow{2}{*}{Avg.}\\ \cline{3-13}
                               &                      & Inver & $\lambda=-20$            & $\lambda=20$               & $\lambda=20$-adv     & Random        & Full    & 0.1              & 0.08             & 0.05              & rp          & rd      \\ \hline
RN50                    & 92.69\%                                       & 1.97\%                                     & 7.30\%                            & 13.35\%                          & 29.92\%                              & 13.34\%                    & 2.71\%                   & 7.25\%                  & 10.51\%                  & 21.26\%                  & 26.46\%                      & 25.12\%                       & 15.72\%               \\
RN50-Adversarial                  & 81.96\%                                       & \textbf{0.66\%}                            & \textbf{4.75\%}                   & 13.62\%                          & 37.87\%                              & 15.25\%                    & 4.87\%                   & 9.62\%                  & 13.94\%                  & 25.51\%                  & 32.51\%                      & 31.96\%                       & 18.99\%               \\
RN50-SIN                & 91.57\%                                       & 2.23\%                                     & 7.61\%                            & 12.19\%                          & 33.16\%                              & 13.58\%                    & 1.68\%                   & 8.30\%                  & 12.60\%                  & 24.23\%                  & 29.16\%                      & 27.24\%                       & 16.98\%               \\
RN50-Debiased           & 93.34\%                                       & 1.43\%                                     & 6.09\%                            & 11.45\%                          & 27.99\%                              & 12.12\%                    & 1.98\%                   & 5.53\%                  & 8.76\%                   & 19.27\%                  & 24.01\%                      & 24.97\%                       & 14.22\%               \\
RN50-Augmix             & 93.50\%                                       & 0.98\%                                     & 6.26\%                            & 8.38\%                           & 30.49\%                              & 12.94\%                    & 1.61\%                   & 6.40\%                  & 9.97\%                   & 21.42\%                  & 27.14\%                      & 22.42\%                       & 14.70\%               \\
RN50-ANT                & 91.87\%                                       & 1.68\%                                     & 6.62\%                            & 11.94\%                          & 35.66\%                              & 15.36\%                    & 1.57\%                   & 7.12\%                  & 10.62\%                  & 21.49\%                  & 26.66\%                      & 25.23\%                       & 16.23\%               \\
RN50-DeepAugment        & 92.88\%                                       & 1.50\%                                     & 6.62\%                            & 12.37\%                          & 32.40\%                              & 13.32\%                    & \textbf{1.36\%}          & 7.27\%                  & 10.62\%                  & 21.28\%                  & 26.28\%                      & 21.29\%                       & 15.28\%               \\
RN50-T                  & \textbf{94.55\%}                              & 1.05\%                                     & 5.65\%                            & \textbf{7.38\%}                  & \textbf{21.89\%}                     & \textbf{10.42\%}           & 2.11\%                   & \textbf{4.74\%}         & \textbf{7.83\%}          & \textbf{17.46\%}         & \textbf{21.12\%}             & \textbf{19.60\%}              & \textbf{11.82\%}     
\\
\toprule
\end{tabular}}
\label{tab:RN50}
\vspace{-2mm}
\end{table*}

\noindent
\textbf{Training strategy.}
a) \textit{Robust trained.}
There have been plenty of studies focusing on the robust training strategy to improve model robustness. To find out whether these works can boost the robustness on our dataset, we further evaluate these state-of-the-art models including SIN~\cite{SIN}, DebiasedCNN~\cite{debiased}, Augmix~\cite{hendrycks2020Augmix}, ANT~\cite{ANT}, DeepAugment~\cite{DeepAugment} and model trained with lots of standard augmentations (RN50-T)~\cite{wightman2021resnet}.
% Fig.~\ref{fig:adv} shows the drop rates of both normal trained and adversarial trained models. 
As shown in Tab.~\ref{tab:RN50}, apart from the RN50-T, while the Augmix model shows the best performance against the background change scenario, the Debiased model holds the best in the object size change scenario. 
% We think the Debiased model benefits from its trained strategy. With the object goes smaller, the texture impact declines. A shape-based model at this situation will still show better robustness against other models. 
What we find unexpectedly is the SIN performance.
The SIN method features the novel data augmentation scheme where ImageNet images are stylized with style transfer as the training data to force the model to rely less on textural cues for classification.
Though the robustness boost is achieved on ImageNet-C~(mCE 69.32\%) compared to its vanilla model~(mCE 76.7\%), it fails to improve the robustness in both object background and size-changing scenarios. 
% For example, the top-1 accuracies for vanilla RN50 and RN50-SIN are 71.09\% and 67.23\% when the object size rate is 0.05 though they share similar accuracy on original ImageNet.
The drops of top-1 accuracy for vanilla RN50 and RN50-SIN are 21.26\% and 24.23\%
% (top-1 accuracies are 71.09\% and 67.23\%) 
respectively, when the object size rate is 0.05, though they share similar accuracy on original ImageNet.
This indicates that existing benchmarks cannot reflect the real robustness in object attribute changing. Therefore, a dataset like ImageNet-E is necessary for comprehensive evaluations on deep models.
% c) Model distilling. 
% d) Data augmentation.
b) \textit{Masked image modeling.}
% Considering that the masked image modeling can reconstruct the whole images with few patches inside the input image,
Considering that masked image modeling has demonstrated impressive results in self-supervised representation learning by recovering corrupted image patches~\cite{bao2021beit}, it may be robust to the attribute changes. Therefore, we choose the Masked AutoEncoder~(MAE)~~\cite{he2022masked} as the training strategy since its objective is recovering images with only 25\% patches.
Specifically, we adopt the MAE training strategy with ViT-B backbone and then finetune it with ImageNet training data.
We find that the robustness is improved. For example, the dropped accuracy decreases from 10.62\% to 9.05\% on average compared to vanilla ViT-B.
% Motivated by the success of MAE, we also test another classical self-supervised learning method----the contrastive learning-based MoCo-V3~\cite{chen2021empirical} method but fail to get a boost. We suspect that the MoCo-V3 pays more attention to the global feature instead of the region of interest since a small change in the background can lead to a high drop rate~(19.29\%) in accuracy. 
% With the great progress in self-supervised learning methods, they have achieved comparable performances in object recognition. Thus we also evaluated some representative approaches including the contrastive learning based MoCo-V3~\cite{chen2021empirical} and reconstruction-based Masked Autoencoders~(MAE)~\cite{he2022masked} for comprehensive studies.
% The results shows that the MAE method is quite robust in our Image-E dataset while MoCo-v3 shows poor performance.
% We think the robustness of MAE benefits from its training strategy, which learns to reconstruct the input image with few patches. Thus simple attribute change does not affect much.

\subsection{Failure case analysis} % overlap analysis, cam pictures
\begin{figure}[!htb]
    \centering
    \includegraphics[height=4cm]{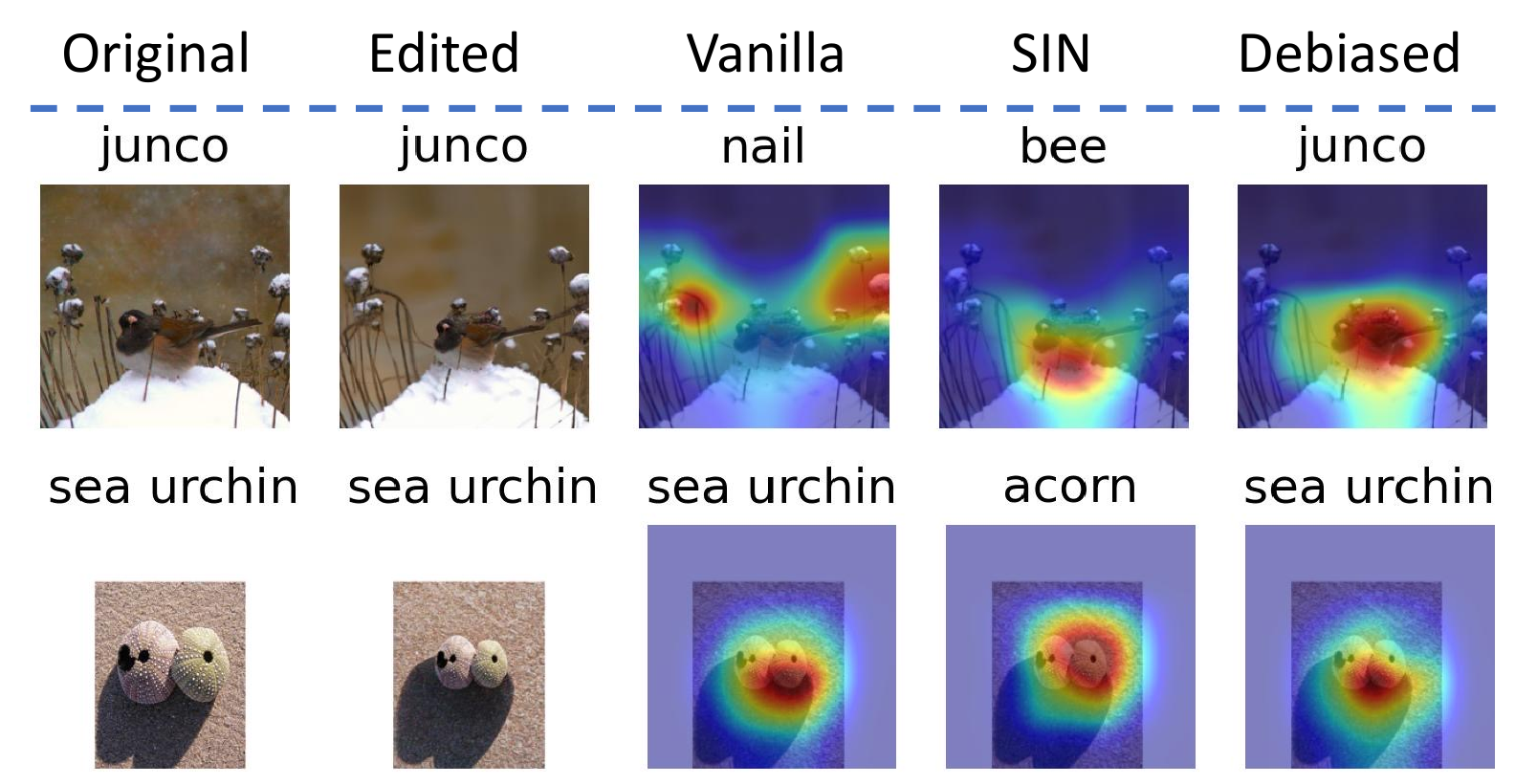}
\caption{Heat maps for explaining which parts of the image dominate the model decision through LayerCAM~\cite{jiang2021layercam}.}
    % \vspace{-5mm}
    \label{fig:cam}
\end{figure}

To explore the reason why some robust trained models may fail, we leverage the LayerCAM~\cite{jiang2021layercam} to generate the heat map for different models including vanilla RN50, RN50+SIN and RN50+Debiased for comprehensive studies. 
As shown in Fig.~\ref{fig:cam}, the heat map of the Debiased model aligns better with the objects in the image than that of the original model. It is interesting to find that the SIN model sometimes makes wrong predictions even with its attention on the main object. We suspect that the SIN model relies too much on the shape. for example, the `sea urchin' looks like the `acron' with the shadow. However, its texture clearly indicates that it is the `sea urchin'. In contrast, the Debiased model which is trained to focus on both the shape and texture can recognize it correctly. More studies can be found in Appendix~\ref{sec:A_bad_case_analysis}.

\subsection{Model repairing}
To validate that the evaluation on ImageNet~(IN)-E can help to provide some insights for model’s applicability and enhancement, we conduct a toy example for model repairing. Previous evaluation shows that the ResNet50 is vulnerable to background changes. Based on this observation, we randomly replace the backgrounds of objects with others during training and get a validation accuracy boost from 77.48\% to 79.00\%. 
% Note that the promotion is not small as we only conduct this operation on 8781 training images since the ImageNet-S only provides 8781 annotated images with object masks in the training data of ImageNet. 
Note that the promotion is not small as only 8781 training images with mask annotations are available in ImageNet.
We also step further to find out if the improved model can get a boost the OOD robustness, as shown in the Tab.~\ref{tab:repair}. It can be observed that with the insights provided by the evaluation on ImageNet-E, one can explore the model’s attribute vulnerabilities and manage to repair the model and get a performance boost accordingly.

\begin{table}[ht]
% \small
\caption{Model repairing results. Top-1 accuracy~(\%) is reported except for IN-C, which is mCE (mean Corruption Error). Higher top-1 accuracy and lower mCE indicate better performance. IN-E reports the average accuracy on ImageNet-E.}
\resizebox{0.45\textwidth}{!}{
\begin{tabular}{c|c|cccccc}
\toprule
Models        & IN    & IN-v2 & IN-A & IN-C$\downarrow$ & IN-R & IN-Sketch & IN-E \\ \hline
RN50          & 77.5 & 65.7  & 6.5  & 68.6 & 39.6 & 27.5      & 83.7 \\
RN50-repaired & \textbf{79.0} & \textbf{67.2} & \textbf{9.4} & \textbf{65.8} & \textbf{40.7} & \textbf{29.4} & \textbf{85.0} 
\\ \toprule
\end{tabular}}
\vspace{-2mm}
\label{tab:repair}
\end{table}

\section{Conclusion and Future work}
In this paper, we put forward an image editing toolkit that can take control of object attributes smoothly. With this tool, we create a new dataset called ImageNet-E that can serve as a general dataset for benchmarking robustness against different object attributes.
Extensive evaluations conducted on different state-of-the-art models show that most models are vulnerable to attribute changes, especially the adversarially trained ones. Meanwhile, other robust trained models can show worse results than vanilla models even when they have achieved a great robustness boost on other robustness benchmarks.
We further discover ways for robustness enhancement from both preprocessing, network designing and training strategies.

\vspace{2mm}
\noindent
\textbf{Limitations and future work.} This paper proposes to edit the object attributes in terms of backgrounds, sizes, positions and directions. Therefore, the annotated mask of the interest object is required, resulting in a limitation of our method.
Besides, since our editing toolkit is developed based on diffusion models, the generalization ability is determined by DDPMs. For example, we find synthesizing high-quality person images is difficult for DDPMs.
Under the consideration of both the annotated mask and data quality, our ImageNet-E is a compact test set.
% In our future work, we would like to extend our method with weakly supervised object localization, thus can help to get rid of the dependency on annotation masks.
In our future work, we would like to explore how to leverage the edited data to enhance the model's performance, including both the validation accuracy and robustness.

\clearpage

%%%%%%%%% REFERENCES
{\small
\bibliographystyle{ieee_fullname}
\bibliography{egbib}
}

\appendix
\section{Details for ImageNet-E}
\label{sec:ImageNet-E}
\begin{figure}[!htb]
    \centering
    \includegraphics[height=5.cm]{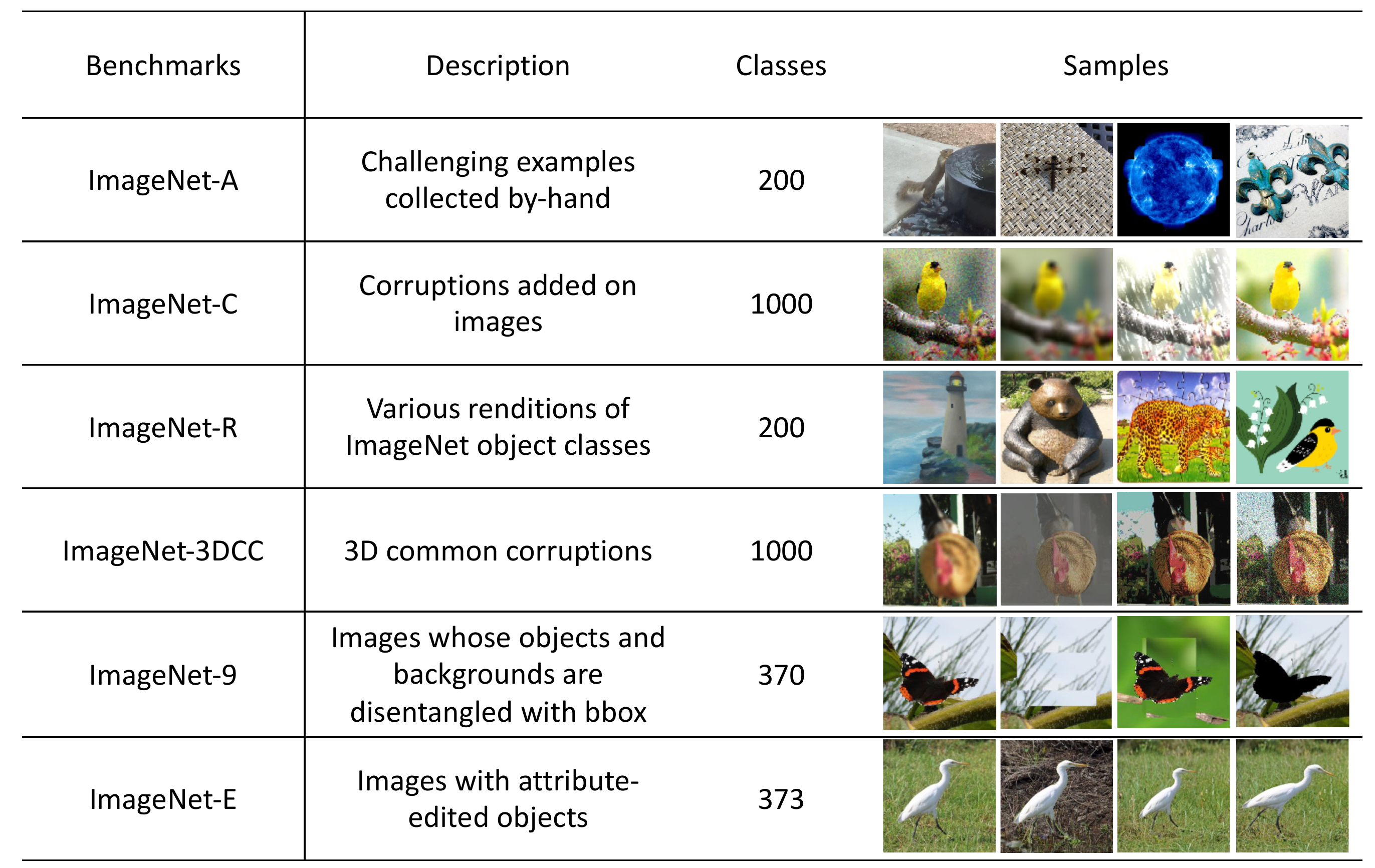}
\caption{Benchmark comparison.}
    % \vspace{-5mm}
    \label{fig:benchmarks}
\end{figure}

\begin{figure*}[t]
    \centering
    \includegraphics[width=\textwidth]{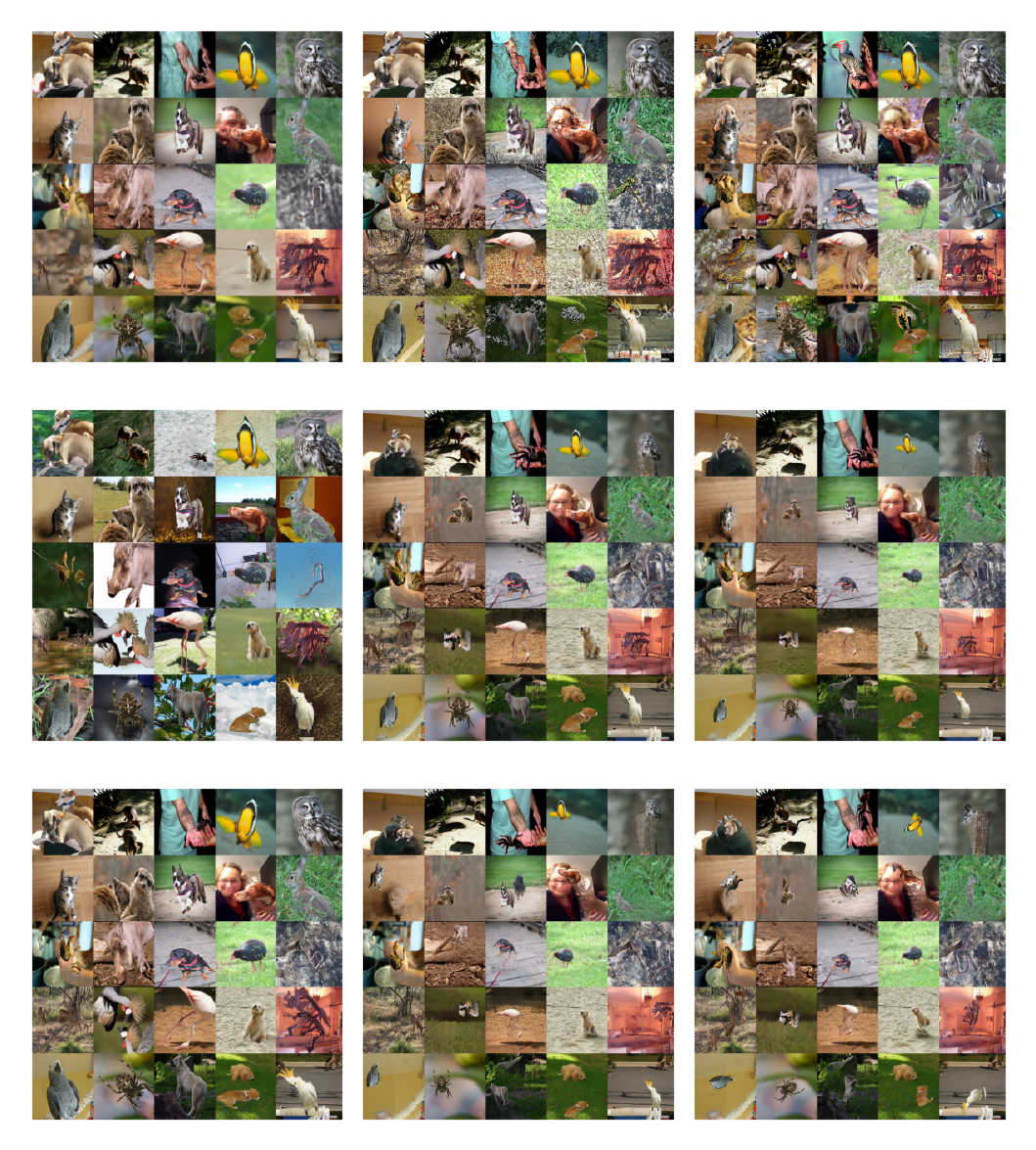}
\caption{Samples from ImageNet-E. From left to right, top to bottom, the images stand for background editing with $\lambda=-20$, $\lambda=20$, $\lambda=20$-adv, randomly shuffled backgrounds, size editing with rate 0.1 and 0.05, randomly rotate, random position, randomly rotate based on images with object pixel rate 0.05 respectively.}
    % \vspace{-5mm}
    \label{fig:examples}
    \vspace{10mm}
\end{figure*}

To guarantee the visual quality of the generated examples, we choose the animal classes from ImageNet since they appear more in nature without messy backgrounds. 
% Classes such as stove and mortarboard are removed. 
Specifically, images whose coarse labels in [fish, shark, bird, salamander, frog, turtle, lizard, crocodile, dinosaur, snake, trilobite, arachnid, ungulate, monotreme, marsupial, coral, mollusk, crustacean, marine mammals, dog, wild dog, cat, wild cat, bear, mongoose, butterfly, echinoderms, rabbit, rodent, hog, ferret, armadillo,primate] are picked. The corresponding coarse labels of each class we refer to can be found in \cite{eshed_novelty_detection}\footnote{https://github.com/noameshed/novelty-detection/blob/master/imagenet\_categories\_synset.csv}.
Finally, our ImageNet-E consists of 373 classes. Since the number of masks provided in ImageNet-S~\cite{ImageNet-S} in these classes is 4352, thus the number of images in each edited kind is 4352. The ImageNet-E contains $11$ kinds of attributes editing, including $5$ kinds of background editing and $4$ kinds of size editing, as well as one kind of position editing and one kind of direction editing. 
Finally, our ImageNet-E contains 47872 images.
Experiments on more images can be found in section~\ref{sec:A_more_data}. The comprehensive comparisons with the state-of-the-art robustness benchmarks are shown in Figure~\ref{fig:benchmarks}. 
In contrast to other benchmarks that investigate new out-of-distribution corruptions or perturbations deep models may encounter, w conduct model debugging with in-distribution data to explore which object attributes a model may be sensitive to.
The examples in ImageNet-E are shown in Figure~\ref{fig:examples}. A demo video for our editing toolkit can be found at this url:\url{https://drive.google.com/file/d/1h5EV3MHPGgkBww9grhlvrl--kSIrD5Lp/view?usp=sharing}. Our code can be found at an anonymous url: \url{https://huggingface.co/spaces/Anonymous-123/ImageNet-Editing}.

\begin{figure}[ht]
    \centering
    \includegraphics[width=0.45\textwidth]{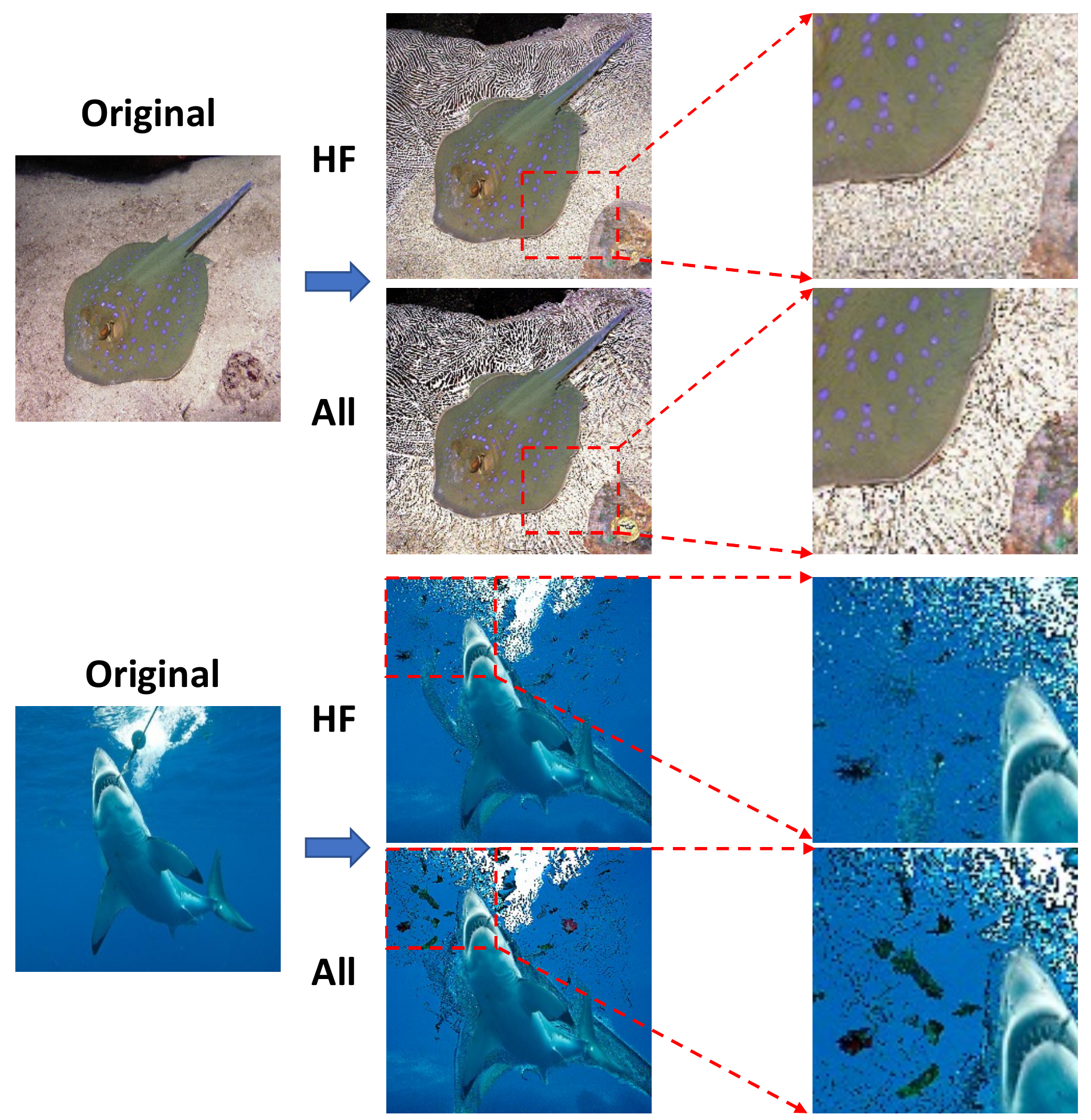}
\caption{Comparisons between the amplitude supervision on high-frequency components~(HF) and amplitude supervision on all frequency components~(All).}
    % \vspace{-5mm}
    \label{fig:HF}
\end{figure}

\section{Background editing}
\label{appendix:bg}
Intuitively, an image with complicated background tends to contain more high-frequency components, such as edges. 
% Higher amplitude indicates greater complexity. 
Therefore, a straight-forward way is to define the background complexity as the amplitude of high-frequency components. However, this operation can result in noisy backgrounds, instead of the ones with complicated textures. Therefore, we directly define complexity as the amplitude of all frequency components. The compared results are shown in Figure~\ref{fig:HF}. It can be observed that the amplitude supervision on high-frequency components tends to make the model generate images with more noise. In contrast, amplitude supervision on all frequency components can help to generate images with texture-complex backgrounds.
To edit the background adversarially, we set $\mathcal{L}_c = \text{CE}(f(\mathbf{x}),y)$ where `CE' is the cross entropy loss. $f$ and $y$ are the classifier and label of $\mathbf{x}$ respectively. We adopt the classifier $f$ from guided-diffusion\footnote{https://github.com/openai/guided-diffusion}.

% \bigskip
% \newpage

\begin{figure*}[t]
    \centering
    \includegraphics[width=\textwidth]{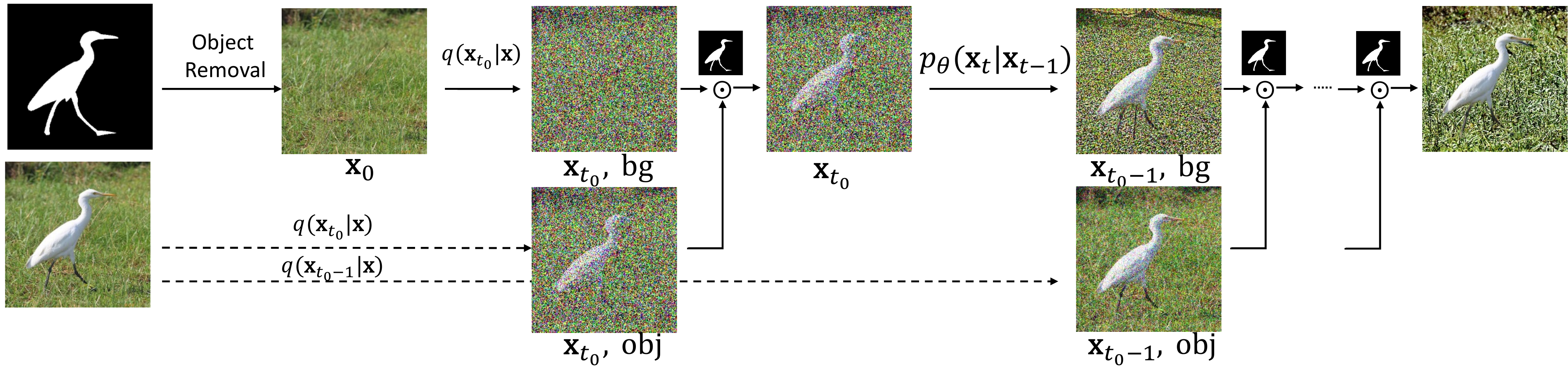}
\caption{Attribute editing with DDPMs. Give an input image and its corresponding object mask, the object is firstly removed with inpainting operation to get the pure background image. Then, we leverage the diffusion process to edit the background image $\mathbf{x}_0$ and object image coherently. $\odot$ denotes the element-wise blending of these two images using the object mask. For background editing, the background complexity objective function is added during the diffusion process (Alg.~\ref{algo:bg}, line $5$). For other object attributes editing, the object image needs to be transformed first (Alg.~\ref{algo_disjdecomp}, line $1$).}
    % \vspace{-5mm}
    \label{fig:DDPM}
\end{figure*}

% \section{Image Editing with Denoising Diffusion Probabilistic Models}
% Previous work has shown that two images can be blended smoothly by separately blending each level of their noisy version~\cite{avrahami2022blended}. The key hypothesis is that at each step during the diffusion process, a noisy latent is projected onto a manifold of natural images noised to a certain level. While blending two noisy images (from the same level) yields a result that likely lies outside the manifold, the next diffusion step projects the result onto the next level manifold, thus ameliorating the incoherence.Inspired by this technique, we follow the same way and fuse the two noisy images~(background and object) to reach the final generated image, as shown in Figure~\ref{fig:DDPM}.

\section{Experimental details}
\label{sec:Appendix_experiment}
\subsection{Details for metrics}
In this paper, we care more about how different attributes impact different models. Therefore, we choose the drop of top-1 accuracy as our evaluation metric. A lower dropped accuracy indicates higher robustness against our attribute changes. The dropped accuracy is defined as:
\begin{equation}
    \text{DA} = \text{acc}_\text{original} - \text{acc}.
\end{equation}
The detailed top-1 accuracy~(Top-1) and dropped accuracy~(DA)on our ImageNet-E are listed in Table~\ref{tab:bg_arch}, Table~\ref{tab:R50-bg} and Table~\ref{tab:obj_size_arch}, Table~\ref{tab:R50-size}. All the experiments are conducted for 5 runs and we report the mean value in the tables.
% RN50-T is the ResNet50 model from timm library which is trained with lots of training strategies~\cite{wightman2021resnet}.

\begin{table*}[ht]
% \color{blue}
\caption{Evaluations under different backgrounds.}
\small
\resizebox{1.0\linewidth}{!}{
\begin{tabular}{c|c|cc|cc|cc|cc|cc}
\toprule
\multirow{2}{*}{Models} & Ori    & \multicolumn{2}{c|}{Inver} & \multicolumn{2}{c|}{$\lambda=-20$} & \multicolumn{2}{c|}{$\lambda=20$} & \multicolumn{2}{c|}{$\lambda=20$-Adv} & \multicolumn{2}{c}{Random} \\ \cline{2-12}
                            & Top-1  & Top-1    & DA                & Top-1     & DA                 & Top-1    & DA                 & Top-1    & DA                  & Top-1      & DA                   \\ \hline\hline
RN50                    & 92.69\%          & 90.72\%          & 1.97\%          & 85.39\%          & 7.30\%          & 79.34\%          & 13.35\%         & 62.77\%           & 29.92\%          & 79.35\%          & 13.34\%         \\
DenseNet121             & 92.10\%          & 90.61\%          & 1.49\%          & 85.81\%          & 6.29\%          & 83.10\%          & 9.00\%          & 62.90\%           & 29.20\%          & 79.67\%          & 12.43\%         \\
EF-B0                   & 92.85\%          & 91.78\%          & 1.07\%          & 85.75\%          & 7.10\%          & 82.14\%          & 10.71\%         & 57.97\%           & 34.88\%          & 77.21\%          & 15.64\%         \\
ResNest50               & 95.38\%          & 93.94\%          & 1.44\%          & 89.05\%          & 6.33\%          & 86.40\%          & 8.98\%          & 68.76\%           & 26.62\%          & 84.10\%          & 11.28\%         \\
ViT-S                   & 94.14\%          & 93.32\%          & \textbf{0.82\%} & 87.72\%          & 6.42\%          & 85.16\%          & 8.98\%          & 63.02\%           & 31.12\%          & 81.08\%          & 13.06\%         \\
Swin-S                  & \textbf{96.21\%} & \textbf{95.08\%} & 1.13\%          & 91.03\%          & 5.18\%          & 88.88\%          & 7.33\%          & 72.71\%           & 23.50\%          & 86.90\%          & 9.31\%          \\
ConvNeXt-T              & 96.07\%          & 94.64\%          & 1.43\%          & \textbf{91.38\%} & \textbf{4.69\%} & \textbf{89.81\%} & \textbf{6.26\%} & \textbf{76.24\%}  & \textbf{19.83\%} & \textbf{88.14\%} & \textbf{7.93\%} \\
\hline\hline                        
RN101                   & 94.00\%          & 91.89\%          & 2.11\%          & 86.95\%          & 7.05\%          & 82.38\%          & 11.62\%         & 64.53\%           & 29.47\%          & 80.43\%          & 13.57\%         \\
DenseNet169             & 92.37\%          & 91.25\%          & 1.12\%          & 86.56\%          & 5.81\%          & 83.94\%          & 8.43\%          & 64.86\%           & 27.51\%          & 80.76\%          & 11.61\%         \\
EF-B3                   & 94.97\%          & 93.10\%          & 1.87\%          & 87.20\%          & 7.77\%          & 86.57\%          & 8.40\%          & 65.07\%           & 29.90\%          & 82.05\%          & 12.92\%         \\
ResNest101              & 95.54\%          & 94.44\%          & 1.10\%          & 89.96\%          & 5.58\%          & 88.89\%          & 6.65\%          & 72.51\%           & 23.03\%          & 85.14\%          & 10.40\%         \\
ViT-B                  & 95.38\%          & 94.55\%          & 0.83\%          & 90.06\%          & 5.32\%          & 86.95\%          & 8.43\%          & 68.78\%           & 26.60\%          & 84.40\%          & 10.98\%         \\
Swin-B                  & 95.96\%          & 95.17\%          & 0.79\%          & 91.50\%          & 4.46\%          & 89.73\%          & 6.23\%          & 74.52\%           & 21.44\%          & 87.71\%          & 8.25\%          \\
ConvNeXt-B              & \textbf{96.42\%} & \textbf{95.73\%} & \textbf{0.69\%} & \textbf{92.67\%} & \textbf{3.75\%} & \textbf{91.56\%} & \textbf{4.86\%} & \textbf{79.93\%}  & \textbf{16.49\%} & \textbf{90.38\%} & \textbf{6.04\%}
\\
\toprule
\end{tabular}}
\label{tab:bg_arch}
\end{table*}

\begin{table*}[ht]
% \color{blue}
\caption{Evaluations with different robust models under different backgrounds.}
\small
\resizebox{1.0\linewidth}{!}{
\begin{tabular}{c|c|cc|cc|cc|cc|cc}
\toprule
\multirow{2}{*}{Models}       & Ori             & \multicolumn{2}{c|}{Inver} & \multicolumn{2}{c|}{$\lambda=-20$}     & \multicolumn{2}{c|}{$\lambda=20$}      & \multicolumn{2}{c|}{$\lambda=20$-Adv} & \multicolumn{2}{c}{Random}   \\ \cline{2-12}
& Top-1  & Top-1    & DA                & Top-1     & DA                 & Top-1    & DA                 & Top-1    & DA                  & Top-1      & DA                   \\ \hline\hline
RN50                    & 92.69\%          & 90.72\%          & 1.97\%          & 85.39\%          & 7.30\%          & 79.34\%          & 13.35\%         & 62.77\%           & 29.92\%          & 79.35\%          & 13.34\%          \\
RN50-A                  & 81.96\%          & 81.30\%          & 0.66\%          & 77.21\%          & 4.75\%          & 68.34\%          & 13.62\%         & 44.09\%           & 37.87\%          & 66.71\%          & 15.25\%          \\
RN50-SIN                & 91.57\%          & 89.34\%          & 2.23\%          & 83.96\%          & 7.61\%          & 79.38\%          & 12.19\%         & 58.41\%           & 33.16\%          & 77.99\%          & 13.58\%          \\
RN50-debiasd            & 93.34\%          & 91.91\%          & 1.43\%          & 87.25\%          & 6.09\%          & 81.89\%          & 11.45\%         & 65.35\%           & 27.99\%          & 81.22\%          & 12.12\%          \\
RN50-Augmix             & 93.50\%          & 92.52\%          & \textbf{0.98\%} & 87.24\%          & 6.26\%          & 85.12\%          & 8.38\%          & 63.01\%           & 30.49\%          & 80.56\%          & 12.94\%          \\
RN50-ANT                & 91.87\%          & 90.19\%          & 1.68\%          & 85.25\%          & 6.62\%          & 79.93\%          & 11.94\%         & 56.21\%           & 35.66\%          & 76.51\%          & 15.36\%          \\
RN50-DeepAugment        & 92.88\%          & 91.38\%          & 1.50\%          & 86.26\%          & 6.62\%          & 80.51\%          & 12.37\%         & 60.48\%           & 32.40\%          & 79.56\%          & 13.32\%          \\
RN50-T                  & \textbf{94.55\%} & \textbf{93.50\%} & 1.05\%          & \textbf{88.90\%} & \textbf{5.65\%} & \textbf{87.17\%} & \textbf{7.38\%} & \textbf{72.66\%}  & \textbf{21.89\%} & \textbf{84.13\%} & \textbf{10.42\%}

\\ \toprule
\end{tabular}}
\label{tab:R50-bg}
\end{table*}

\begin{table*}[ht]
% \color{blue}
\caption{Evaluations under different object sizes.}
\centering
\small
\resizebox{1.0\linewidth}{!}{
\begin{tabular}{c|c|cc|cc|cc|cc|cc|cc}
\toprule
\multirow{2}{*}{Models} & Ori     & \multicolumn{2}{c|}{Full}   & \multicolumn{2}{c|}{0.10}   & \multicolumn{2}{c|}{0.08}   & \multicolumn{2}{c|}{0.05}   & \multicolumn{2}{c|}{0.05-rp} & \multicolumn{2}{c}{rd} \\ \cline{2-14}
& Top-1  & Top-1    & DA    & Top-1     & DA              & Top-1     & DA                 & Top-1    & DA                 & Top-1    & DA          & Top-1    & DA            \\ \hline\hline
RN50                   & 92.69\%          & 89.98\%          & 2.71\%          & 85.44\%          & 7.25\%          & 82.18\%          & 10.51\%         & 71.43\%          & 21.26\%          & 66.23\%          & 26.46\%          & 67.57\%          & 25.12\%          \\
DenseNet121            & 92.10\%          & 88.60\%          & 3.50\%          & 85.10\%          & 7.00\%          & 81.42\%          & 10.68\%         & 70.55\%          & 21.55\%          & 65.57\%          & 26.53\%          & 68.46\%          & 23.64\%          \\
EF-B0                  & 92.85\%          & 89.82\%          & 3.03\%          & 84.85\%          & 8.00\%          & 81.28\%          & 11.57\%         & 69.57\%          & 23.28\%          & 64.94\%          & 27.91\%          & 73.74\%          & 19.11\%          \\
ResNest50              & 95.38\%          & 92.85\%          & 2.53\%          & 90.11\%          & 5.27\%          & 87.37\%          & 8.01\%          & 77.35\%          & 18.03\%          & 74.01\%          & 21.37\%          & 78.06\%          & 17.32\%          \\
ViT-S                 & 94.14\%          & 93.34\%          & 0.80\%          & 88.77\%          & 5.37\%          & 85.55\%          & 8.59\%          & 76.77\%          & 17.37\%          & 71.28\%          & 22.86\%          & 77.01\%          & 17.13\%          \\
% ViT-S                  & 94.74\%          & 93.52\%          & 1.22\%          & 87.64\%          & 7.10\%          & 84.10\%          & 10.64\%         & 74.45\%          & 20.29\%          & 69.66\%          & 25.08\%          & 77.52\%          & 17.22\%          \\
Swin-S                 & \textbf{96.21\%} & \textbf{94.94\%} & \textbf{1.27\%} & 92.00\%          & 4.21\%          & 89.92\%          & 6.29\%          & 82.05\%          & 14.16\%          & 78.86\%          & 17.35\%          & \textbf{82.79\%} & \textbf{13.42\%} \\
ConvNeXt-T             & 96.07\%          & 94.32\%          & 1.75\%          & \textbf{92.79\%} & \textbf{3.28\%} & \textbf{90.89\%} & \textbf{5.18\%} & \textbf{83.31\%} & \textbf{12.76\%} & \textbf{80.36\%} & \textbf{15.71\%} & 80.29\%          & 15.78\%          \\
                       \hline\hline
RN101                  & 94.00\%          & 91.43\%          & 2.57\%          & 87.19\%          & 6.81\%          & 83.88\%          & 10.12\%         & 73.35\%          & 20.65\%          & 68.15\%          & 25.85\%          & 69.58\%          & 24.42\%          \\
DenseNet169            & 92.37\%          & 90.12\%          & 2.25\%          & 85.47\%          & 6.90\%          & 81.96\%          & 10.41\%         & 71.78\%          & 20.59\%          & 67.44\%          & 24.93\%          & 71.69\%          & 20.68\%          \\
EF-B3                  & 94.97\%          & 93.61\%          & 1.36\%          & 88.17\%          & 6.80\%          & 84.81\%          & 10.16\%         & 73.61\%          & 21.36\%          & 69.99\%          & 24.98\%          & 77.73\%          & 17.24\%          \\
ResNest101             & 95.54\%          & 94.19\%          & 1.35\%          & 91.57\%          & 3.97\%          & 89.01\%          & 6.53\%          & 80.10\%          & 15.44\%          & 76.43\%          & 19.11\%          & 81.23\%          & 14.31\%          \\
ViT-B                 & 95.38\%          & 94.76\%          & \textbf{0.62\%}          & 91.38\%          & 4.00\%          & 89.08\%          & 6.30\%          & 80.87\%          & 14.51\%          & 76.56\%          & 18.82\%          & 80.43\%          & 14.95\%          \\
% ViT-B                  & 95.66\%          & 95.20\%          & \textbf{0.46\%} & 90.97\%          & 4.69\%          & 89.09\%          & 6.57\%          & 79.98\%          & 15.68\%          & 76.19\%          & 19.47\%          & \textbf{84.22\%} & \textbf{11.44\%} \\
Swin-B                 & 95.96\%          & 94.97\%          & 0.99\%          & 92.80\%          & 3.16\%          & 90.92\%          & 5.04\%          & 83.62\%          & 12.34\%          & 80.58\%          & 15.38\%          & 83.36\%          & \textbf{12.60\%}          \\
ConvNeXt-B             & \textbf{96.42\%} & \textbf{95.43\%} & 0.99\%          & \textbf{94.17\%} & \textbf{2.25\%} & \textbf{93.06\%} & \textbf{3.36\%} & \textbf{86.95\%} & \textbf{9.47\%}  & \textbf{84.02\%} & \textbf{12.40\%} & \textbf{83.41\%}          & 13.01\%         
\\
\toprule     
\end{tabular}}
\label{tab:obj_size_arch}
\end{table*}

\begin{table*}[t]
% \color{blue}
\caption{Evaluations with different robust models under different object sizes.}
\small
\resizebox{1.0\linewidth}{!}{
% \begin{tabular}{c|c|cc|cc|cc|cc|cc}
% \toprule
% \multirow{2}{*}{Object} & Ori   & \multicolumn{2}{c|}{Full}  & \multicolumn{2}{c|}{0.10}   & \multicolumn{2}{c|}{0.08}   & \multicolumn{2}{c|}{0.05}   & \multicolumn{2}{c}{0.05-rp} \\ \cline{2-14}
\begin{tabular}{c|c|cc|cc|cc|cc|cc|cc}
\toprule
\multirow{2}{*}{Models} & Ori     & \multicolumn{2}{c|}{Full}   & \multicolumn{2}{c|}{0.10}   & \multicolumn{2}{c|}{0.08}   & \multicolumn{2}{c|}{0.05}   & \multicolumn{2}{c|}{0.05-rp} & \multicolumn{2}{c}{rd} \\ \cline{2-14}
& Top-1  & Top-1    & DA    & Top-1     & DA              & Top-1     & DA                 & Top-1    & DA                 & Top-1    & DA          & Top-1    & DA            \\ \hline\hline

RN50                   & 92.69\%          & 89.98\%          & 2.71\%          & 85.44\%          & 7.25\%          & 82.18\%          & 10.51\%         & 71.43\%          & 21.26\%          & 66.23\%          & 26.46\%          & 67.57\%          & 25.12\%          \\
RN50-A                 & 81.96\%          & 77.09\%          & 4.87\%          & 72.34\%          & 9.62\%          & 68.02\%          & 13.94\%         & 56.45\%          & 25.51\%          & 49.45\%          & 32.51\%          & 50.00\%          & 31.96\%          \\
RN50-SIN               & 91.57\%          & 89.89\%          & 1.68\%          & 83.27\%          & 8.30\%          & 78.97\%          & 12.60\%         & 67.34\%          & 24.23\%          & 62.41\%          & 29.16\%          & 64.33\%          & 27.24\%          \\
RN50-debiasd           & 93.34\%          & 91.36\%          & 1.98\%          & 87.81\%          & 5.53\%          & 84.58\%          & 8.76\%          & 74.07\%          & 19.27\%          & 69.33\%          & 24.01\%          & 68.37\%          & 24.97\%          \\
RN50-Augmix            & 93.50\%          & 91.89\%          & 1.61\%          & 87.10\%          & 6.40\%          & 83.53\%          & 9.97\%          & 72.08\%          & 21.42\%          & 66.36\%          & 27.14\%          & 71.08\%          & 22.42\%          \\
RN50-ANT               & 91.87\%          & 90.30\%          & 1.57\%          & 84.75\%          & 7.12\%          & 81.25\%          & 10.62\%         & 70.38\%          & 21.49\%          & 65.21\%          & 26.66\%          & 66.64\%          & 25.23\%          \\
RN50-DeepAugment       & 92.88\%          & 91.52\%          & \textbf{1.36\%} & 85.61\%          & 7.27\%          & 82.26\%          & 10.62\%         & 71.60\%          & 21.28\%          & 66.60\%          & 26.28\%          & 71.59\%          & 21.29\%          \\
RN50-T                 & \textbf{94.55\%} & \textbf{92.44\%} & 2.11\%          & \textbf{89.81\%} & \textbf{4.74\%} & \textbf{86.72\%} & \textbf{7.83\%} & \textbf{77.09\%} & \textbf{17.46\%} & \textbf{73.43\%} & \textbf{21.12\%} & \textbf{74.95\%} & \textbf{19.60\%}
\\

\toprule
\end{tabular}}
\label{tab:R50-size}
\end{table*}

% \subsection{The whole pipeline}
% Our whole pipeline is shown in Fig.~\ref{fig:DDPM}. Take the background editing for example, given an input image with the corresponding object mask, we first run object removal process with the public available algorithm TFill to the the clean background $x_0$.  During each step, the 

% \clearpage

\subsection{Classes whose accuracy drops the greatest}
\begin{figure*}[ht]
    \centering
    \includegraphics[height=7cm]{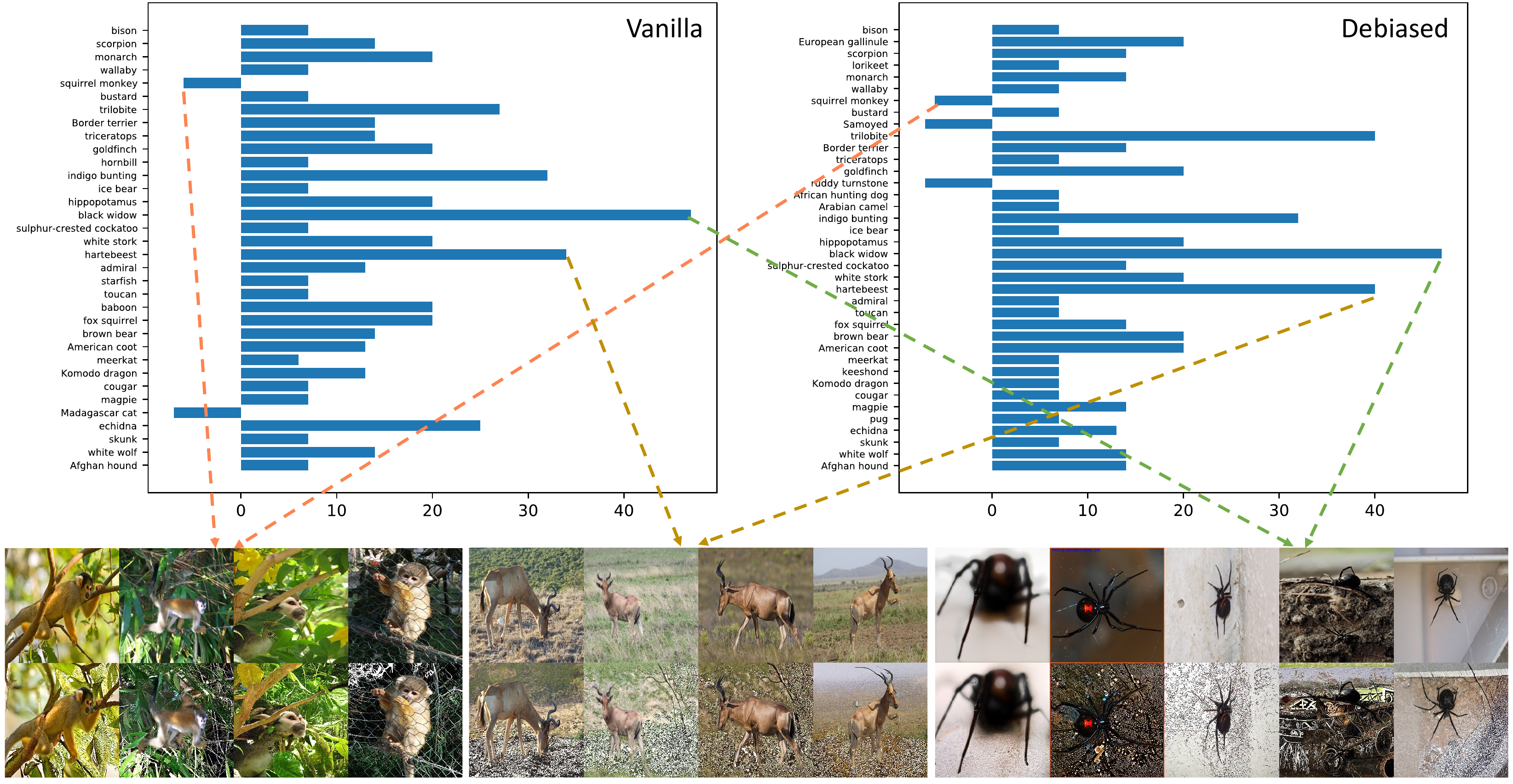}
    \caption{Dropped accuracy~(\%) in each class. Classes whose number of images is less than 15 or dropped accuracy is zero are removed.}
    % \vspace{-5mm}
    \label{fig:cls}
\end{figure*}

To find out which class gets the worst robustness against attribute changes, we plot the dropped accuracy in Figure~\ref{fig:cls}. The evaluated models are vanilla RN50 and its Debiased model. It can be observed that objects that have tentacles with simple backgrounds are more easily to be attacked. For example, the dropped accuracy of the `black widow' class reaches 47\% for both vanilla and Debiased models. In contrast, the impact is smaller for images with complicated backgrounds such as pictures from `squirrel monkey'.

\subsection{Experiments on more data}
\label{sec:A_more_data}
To explore the model robustness against object attributes on large-scale datasets, we step further to conduct the image editing on all the images in the ImageNet-S validation set. Finally, the edited dataset ImageNet-E-L shares the same size as ImageNet-S, which consists of 919 classes and 10919 images. We conduct both background editing and size editing to them. The evaluation results are shown in Table~\ref{tab:more_data}. The same conclusion can also be observed. For instance, most models show vulnerability against attribute changing since the average dropped accuracies reach 12.22\% and 22.21\% in background and size changes respectively. When the model gets larger, the robustness is improved. The consistency implies that using our ImageNet-E can already reflect the model robustness against object attribute changes.

\begin{table*}[ht]
\caption{Evaluations with more data.}
\resizebox{1.0\linewidth}{!}{
\begin{tabular}{c|c|cc|cc|c|c|cc|cc}
\toprule
\multirow{2}{*}{Models} & Original        & \multicolumn{2}{c|}{Background}    & \multicolumn{2}{c|}{Size-0.05}      & \multirow{2}{*}{Models} & Original        & \multicolumn{2}{c|}{Background}    & \multicolumn{2}{c}{Size-0.05}      \\ \cline{2-6} \cline{8-12}
                        & Top-1             & Top-1             & DA              & Top-1             & DA               &                         & Top-1             & Top-1             & DA              & Top-1             & DA               \\ \hline\hline
DenseNet121                & 86.60\%          & 74.73\%          & 11.87\%         & 61.48\%          & 25.12\%          & DenseNet169                & 87.66\%          & 76.26\%          & 11.40\%         & 63.57\%          & 24.09\%          \\
RN50                    & 88.12\%          & 71.64\%          & 16.48\%         & 63.13\%          & 24.99\%          & RN101                   & 89.52\%          & 75.33\%          & 14.19\%         & 65.11\%          & 24.41\%          \\
EF-B0                    & 88.54\%          & 75.64\%          & 12.90\%         & 62.16\%          & 26.38\%          & EF-B3                    & 92.12\%          & 80.81\%          & 11.31\%         & 66.18\%          & 25.96\%          \\
ResNest50               & 92.12\%          & 80.61\%          & 11.51\%         & 70.05\%          & 22.07\%          & ResNest101              & 92.78\%          & 83.46\%          & 9.32\%         & 72.67\%          & 20.11\%          \\
ViT-S                   & 92.15\%          & 78.94\%          & 13.21\%         & 69.30\%          & 22.85\%          & ViT-B                & \textbf{94.12\%} & 83.04\%          & 11.08\%         & 75.65\%          & 18.47\%          \\
Swin-S                  & \textbf{93.11\%} & 82.98\%          & 10.13\%         & 75.36\%          & 17.75\%          & Swin-B                  & 93.18\%          & 84.11\%          & 9.07\%          & 76.99\%          & 16.19\%          \\
ConvNeXt-T              & 92.75\%          & \textbf{84.00\%} & \textbf{9.43\%} & \textbf{76.41\%} & \textbf{16.34\%} & ConvNeXt-B              & 94.05\%          & \textbf{86.41\%} & \textbf{7.64\%} & \textbf{80.34\%} & \textbf{13.71\%} \\ \toprule
\end{tabular}}
\label{tab:more_data}
\end{table*}

\subsection{Bad case analysis}
\label{sec:A_bad_case_analysis}
\begin{figure*}[ht]
    \centering
    \includegraphics[height=7.5cm]{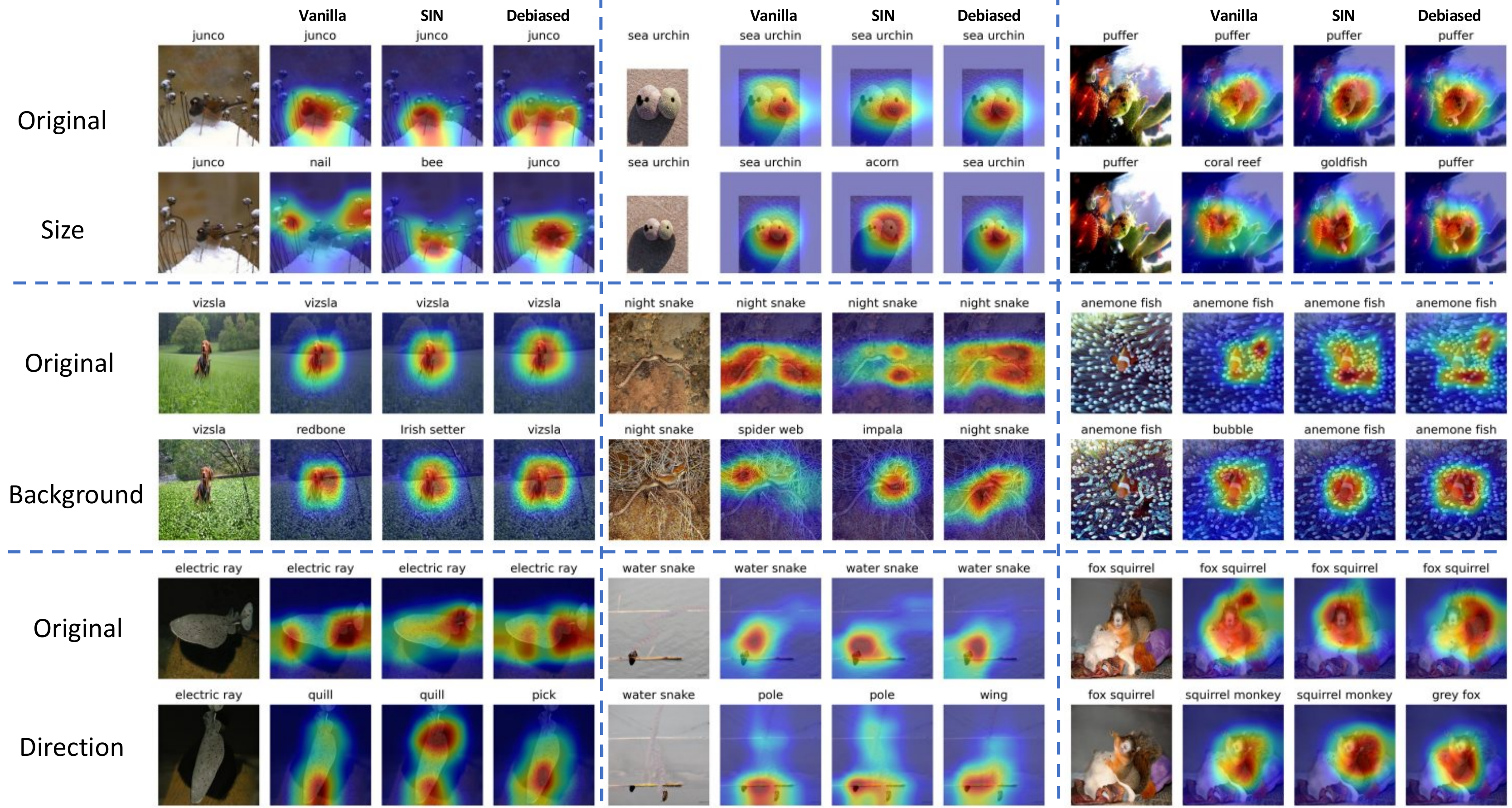}
\caption{The heat map comparisons between original images and edited ones.}
    % \vspace{-5mm}
    \label{fig:bad_case_all}
\end{figure*}

To make a comprehensive study of how the model behaves, we step further to make a comparison of the heat maps of the originals and edited ones. We choose the images that are recognized correctly at first but misclassified after editing. All the attributes editing including background, size, directions are explored. The heat maps are visualized in Figure~\ref{fig:bad_case_all}. It can be observed that compared to the SIN and Debiased models, the vanilla RN50 is more likely to lose its focus on the interest area, especially in the size change scenario. For example, in the second row, as it puts his focus on the background, it returns a result with the `nail' label. The same fashion is also observed in the background change scenario. The predicted label of `night snake' turns into `spider web' as the complex background has attracted its attention. 
In contrast, the SIN and Debiased models have robust attention mechanisms. 
The quantitative results in Table~\ref{tab:R50-bg} also validate this. The dropped accuracy of RN50~(13.35\%) is higher than SIN~(12.19\%) and Debiased~(11.45\%) even though the original accuracy of SIN~(0.9157) is lower than vanilla RN50~(0.9269).
However, the SIN also has its weakness. We find that though the SIN pays attention to the desired region, it can also make wrong predictions. As shown in the second row of Figure~\ref{fig:bad_case_all}, when the object size gets smaller, the shape-based SIN model tends to make wrong predictions, \textit{e.g.}, mistaking the `sea urchin' as 'acorn' due to the lack of texture analysis. As a result, the dropped accuracy in the size change scenario is 24.23\% for SIN, even lower than vanilla RN50, whose dropped accuracy is 21.26\%.
On the contrary, the Debiased model can recognize it correctly, profiting from its shape and texture-biased module.
From the above observation, we can conclude that the texture matters in the small object scenario.

\subsection{Details for robustness enhancements}
\label{sec:robust_enhance}
\textbf{Network design----self-attention-like architecture.} 
The results in Table~\ref{tab:arch} show that most vision transformers show better robustness than CNNs in our scenario. Previous study has shown that the self-attention-like architecture may be the key to robustness boost~\cite{bai2021transformers}. Therefore, to ablate whether incorporating this module can help attribute robustness generalization, we create a hybrid architecture~(RN50d-hybrid) by directly feeding the output of res\_3 block in RN50d into ViT-S as the input feature. The results are shown in Table~\ref{tab:self-att}. 
% As we can find that the added module can help to boost the robustness against size changes, but fails in promoting the robustness on background changes. 
As we can find that while the added module maintains the robustness on background changes, it can help to boost the robustness against size changes.
Moreover, the RN50-hybrid can also boost the overall performance compared to ViT-S. 
% It seems that the deeper convolutional modules or greater receptive fields can help to capture a more robust feature.

\begin{table*}[t]
% \color{blue}
\caption{Ablation study of the self-attention-like architecture.}
\resizebox{1.0\linewidth}{!}{
\begin{tabular}{c|c|ccccc|cccc|c|c|c}
\toprule
\multirow{2}{*}{Models} & \multirow{2}{*}{Ori} & \multicolumn{5}{c|}{Background changes}                                                   & \multicolumn{4}{c|}{Size changes}                                  & Position   & Direction    & \multirow{2}{*}{Avg.}   \\ \cline{3-13}
                               &                      & Inver & $\lambda=-20$            & $\lambda=20$            & $\lambda=20$-Adv           & Random            & Full    & 0.1              & 0.08             & 0.05              & rp   & rd &             \\ \hline\hline
% RN50d            & 0.9375          & 1.31\%          & \textbf{5.12\%}          & \textbf{6.91\%}          & 12.37\%          & \textbf{20.68\%}          & 2.97\%          & 4.65\%          & 7.54\%          & 18.08\%          & 21.85\%             & 20.60\%             \\
% ViT-S            & 0.9474          & 1.75\%          & 7.73\%          & 11.23\%         & 19.21\%          & 33.96\%          & \textbf{1.29\%}          & 7.50\%          & 11.23\%         & 21.42\%          & 26.48\%             & 18.18\%             \\ \hline
% R50d-hybrid      & \textbf{0.9540}          & \textbf{1.09\%}          & 5.92\%          & 7.51\%          & \textbf{10.95\%}          & 22.57\%          & 1.42\%          & \textbf{3.70\%}          & \textbf{6.21\%}          & \textbf{14.59\%}          & \textbf{18.06\%}             & \textbf{14.80\%}            \\
R50d                    & 93.77\%                                       & 1.23\%                                     & \textbf{4.80\%}                   & \textbf{6.48\%}                  & \textbf{19.39\%}                     & \textbf{8.28\%}            & 2.82\%                   & 4.36\%                  & 7.07\%                   & 16.95\%                  & 20.49\%                      & 19.31\%                       & 11.00\%               \\
ViT-S                   & 94.74\%                                       & 1.66\%                                     & 7.32\%                            & 10.64\%                          & 32.17\%                              & 14.39\%                    & \textbf{1.22\%}                   & 7.10\%                  & 10.64\%                  & 20.29\%                  & 25.08\%                      & 17.22\%                       & 14.61\%               \\
R50-hybrid              & \textbf{95.40\%}                              & \textbf{1.04\%}                            & 5.64\%                            & 7.16\%                           & 21.54\%                              & 9.19\%                     & 1.37\%          & \textbf{3.53\%}         & \textbf{5.92\%}          & \textbf{13.92\%}         & \textbf{17.23\%}             & \textbf{14.12\%}              & \textbf{9.96\%}       \\
\toprule
\end{tabular}}
\label{tab:self-att}
\end{table*}

\textbf{Training strategy----Masked image modeling.}
Considering that masked image modeling has demonstrated impressive results in self-supervised representation learning by recovering corrupted image patches~\cite{bao2021beit}, it may be robust to the attribute changes. Thus, we test the Masked AutoEncoder~(MAE)~\cite{mae} and SimMIM~\cite{xie2022simmim} training strategy based on ViT-B backbone. As shown in Table~\ref{tab:mae}, the dropped accuracies decrease a lot compared to vanilla ViT-B, validating the effectiveness of the masked image modeling strategy. 
Motivated by this success, we also test another kind of self-supervised-learning strategy. To be specific, we choose the representative method MoCo-V3~\cite{chen2021empirical} in the contrastive learning family. After the end-to-end finetuning, it achieves top-1 83.0\% accuracy on ImageNet.
It can also improve the attribute robustness when compared to the vanilla ViT-B, showing the effectiveness of contrastive learning.
\begin{table*}[t]
% \color{blue}
\caption{Ablation study of the self-supervised models. All the compared models are end-to-end finetuned on ImageNet except for ViT-B, which is supervised trained from the early start.}
\resizebox{1.0\linewidth}{!}{
\begin{tabular}{c|c|ccccc|cccc|c|c|c}
\toprule
\multirow{2}{*}{Models} & \multirow{2}{*}{Ori} &  \multicolumn{5}{c|}{Background changes}                                                   & \multicolumn{4}{c|}{Size changes}                                  & Position   & Direction   & \multirow{2}{*}{Avg.}    \\ \cline{3-13}
                               &                      & Inver & $\lambda=-20$            & $\lambda=20$            & $\lambda=20$-Adv           & Random            & Full    & 0.1              & 0.08             & 0.05              & rp   & rd      &       \\ \hline\hline

ViT-B                   & 95.38\%                              & 0.83\%                                     & 5.32\%                            & 8.43\%                           & 26.60\%                              & 10.98\%                    & 0.62\%                   & 4.00\%                  & 6.30\%                   & 14.51\%                  & 18.82\%                      & 14.95\%                       & 11.05\%               \\
CLIP\_finetune          & 93.68\%                                       & 2.17\%                                     & 9.82\%                            & 11.83\%                          & 38.33\%                              & 18.19\%                    & 9.06\%                   & 9.25\%                  & 12.67\%                  & 23.32\%                  & 28.56\%                      & 22.00\%                       & 18.30\%               \\
MoCo-v3                 & 95.70\%                                       & \textbf{0.55\%}                            & 4.91\%                            & 7.33\%                           & 24.33\%                              & 9.92\%                     & 0.92\%                   & 3.76\%                  & 5.62\%                   & 13.61\%                  & 17.85\%                      & 15.20\%                       & 10.35\%               \\
MAE-ViT-B               & 96.12\%                                       & 0.78\%                                     & \textbf{4.77\%}                   & \textbf{6.21\%}                  & \textbf{21.09\%}                     & \textbf{8.18\%}            & \textbf{0.78\%}          & \textbf{3.01\%}         & \textbf{4.86\%}          & \textbf{12.10\%}         & \textbf{15.47\%}             & 14.00\%                       & \textbf{9.05\%}       \\
SimMIM                  & \textbf{96.14\%}                                       & 0.75\%                                     & 5.13\%                            & 6.76\%                           & 23.58\%                              & 9.33\%                     & 0.97\%                   & 3.22\%                  & 5.33\%                   & 13.18\%                  & 17.12\%                      & \textbf{13.62\%}              & 9.82\%                \\

\toprule
\end{tabular}}
\label{tab:mae}
\end{table*}

\subsection{Hardware}
Our experiments are implemented by PyTorch~\cite{paszke2019pytorch} and runs on RTX-3090TI.

\begin{figure}[ht]
    \centering
    \includegraphics[width=0.45\textwidth]{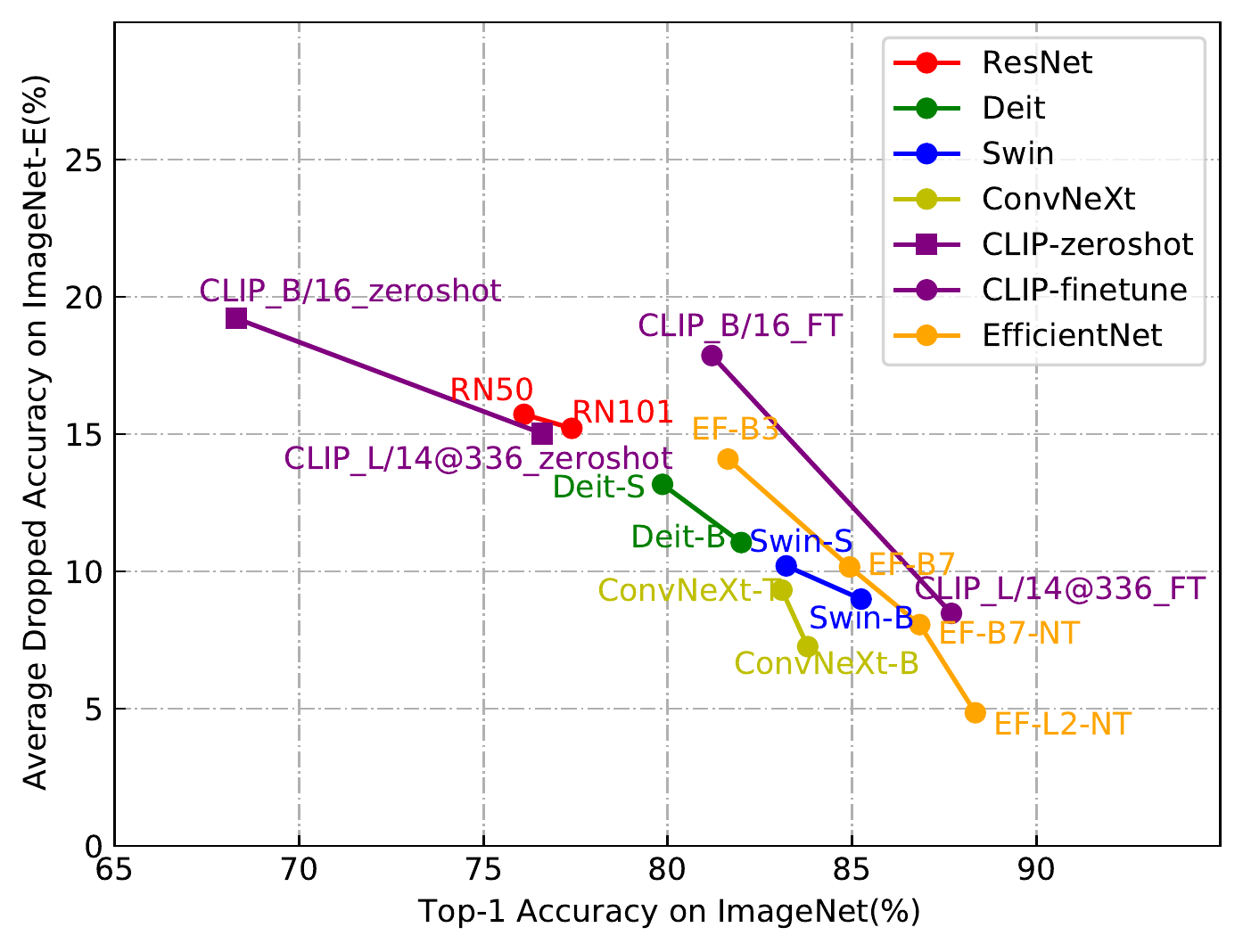}
\caption{The average accuracy drop of different models. The $x$-axis is the model's top-1 accuracy on ImageNet.}
    % \vspace{-5mm}
    \label{fig:top1_DA}
\end{figure}

\section{Further exploration on backgrounds}
Motivated by the models' vulnerability against background changes, especially for those complicated backgrounds. Apart from randomly picking the backgrounds from the ImageNet dataset as final backgrounds~(random\_bg), we also collect background templates with abundant textures, including leopard, eight diagrams, checker and stripe to explore the performance on out-of-distribution backgrounds. The evaluation results are shown in Table~\ref{tab:bgs}. 
% Note that the FID for images generated with random backgrounds is 15.57, indicating that these images are approximately distributed in the ImageNet distribution, thus can be viewed as in-distribution data. 
It can be observed that the background changes can lead to a 13.34\% accuracy drop. When the background is set to be a leopard or other images, the dropped accuracy can even reach 35.52\%. Sometimes the robust models even show worse robustness. For example, when the background is eight diagrams, all the robust models show worse results than the vanilla RN50, which is quite unexpected. 
To comprehend the behaviour behind it, we visualize the heat maps of the different models in Figure~\ref{fig:benchmarks}.
An interesting finding is that deep models tend to make decisions with dependency on the backgrounds, especially when the background is complicated and can attract some attention. 
For example, when the background is the eight diagrams, the SIN takes the goldfish as a dishwasher. We suspect it has mistaken the background as dishes. In the same fashion, the Debiased model and ANT take the `sea slug' with eight diagrams as a `shopping basket', which seems to make sense since the `sea slug' looks like a vegetable.

\begin{table}[ht]
% \color{blue}
\centering
\caption{Evaluations on different robustness benchmarks. All results are top-1 accuracies(\%) on corresponding datasets except for ImageNet-C, which is mCE~(mean Corruption Error). Higher top-1 accuracy and lower mCE indicate better performance.}
\resizebox{1.0\linewidth}{!}{
\begin{tabular}{c|c|ccccc|c}
\toprule
Models         & IN            & IN-V2         & IN-A          & IN-C          & IN-R          & IN-Sketch     & IN-E \\ \hline
CLIP-zero-shot & 68.3          & 61.9          & \textbf{50.1} & \textbf{43.1} & \textbf{77.6} & \textbf{48.3} & 62.1  \\
CLIP-FT        & \textbf{81.2} & \textbf{70.7} & 35.3          & 47..9         & 65.0          & 44.9          & \textbf{77.2}       \\ \toprule
\end{tabular}}
\label{tab:CLIP_OOD}
\end{table}

\begin{figure*}[t]
    \centering
    \includegraphics[width=\textwidth]{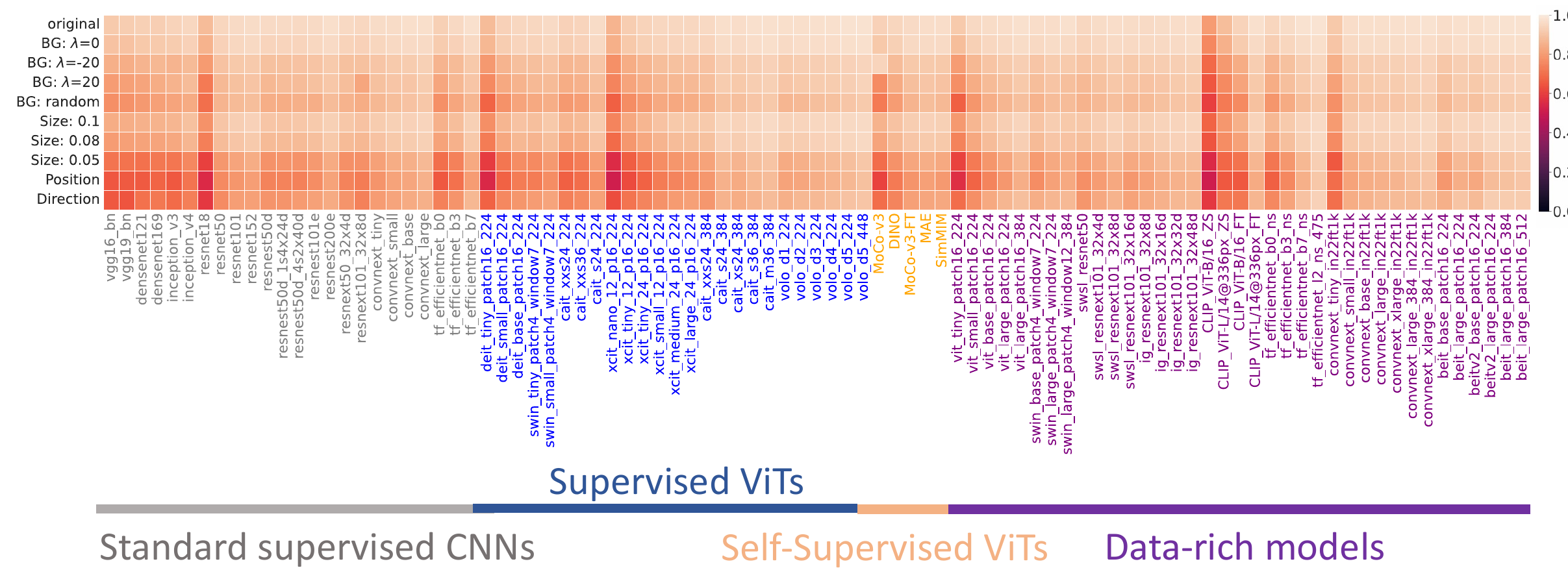}
\caption{The top-1 accuracy performance under different editing scenarios of 91 state-of-the-art models.}
    % \vspace{-5mm}
    \label{fig:91}
\end{figure*}

\begin{table*}[ht]
% \color{blue}
\caption{Evaluation of images generated with different backgrounds. 
% The red ones in each row indicate the background with the worst performance of the corresponding models.
}
\resizebox{1.0\linewidth}{!}{
\begin{tabular}{c|c|cc|cc|cc|cc|cc}
\toprule
\multirow{2}{*}{Models} & \multirow{2}{*}{Original} & \multicolumn{2}{c|}{Random\_bg}      & \multicolumn{2}{c|}{Leopard}         & \multicolumn{2}{c|}{Eight diagrams}  & \multicolumn{2}{c|}{Checker}          & \multicolumn{2}{c}{Stripe}          \\ \cline{3-12}
                        &                           & Top-1            & DA               & Top-1            & DA               & Top-1            & DA               & Top-1             & DA               & Top-1            & DA               \\ \hline\hline
RN50                    & 92.69\%          & 79.35\%          & 13.34\%          & 57.17\%          & 35.52\%          & 64.32\%          & 28.37\%          & 65.13\%           & 27.56\%          & 62.90\%          & 29.79\%          \\
RN50-A                  & 81.96\%          & 66.71\%          & 15.25\%          & 25.05\%          & 56.91\%          & 37.21\%          & 44.75\%          & 32.47\%           & 49.49\%          & 46.96\%          & 35.00\%          \\
RN50-SIN                & 91.57\%          & 77.99\%          & 13.58\%          & 62.74\%          & 28.83\%          & 48.74\%          & 42.83\%          & 51.15\%           & 40.42\%          & 52.65\%          & 38.92\%          \\
RN50-debiasd            & 93.34\%          & 81.22\%          & 12.12\%          & 68.58\%          & 24.76\%          & 62.68\%          & 30.66\%          & 67.10\%           & 26.24\%          & 63.16\%          & 30.18\%          \\
RN50-Augmix             & 93.50\%          & 80.56\%          & 12.94\%          & 57.35\%          & 36.15\%          & 56.20\%          & 37.30\%          & 68.78\%           & 24.72\%          & 65.68\%          & 27.82\%          \\
RN50-ANT                & 91.87\%          & 76.51\%          & 15.36\%          & 58.11\%          & 33.76\%          & 59.04\%          & 32.83\%          & 51.91\%           & 39.96\%          & 54.69\%          & 37.18\%          \\
RN50-DeepAugment        & 92.88\%          & 79.56\%          & 13.32\%          & 62.83\%          & 30.05\%          & 57.71\%          & 35.17\%          & 59.46\%           & 33.42\%          & 61.80\%          & 31.08\%          \\
R50-T                   & \textbf{94.55\%} & \textbf{84.13\%} & \textbf{10.42\%} & \textbf{72.93\%} & \textbf{21.62\%} & \textbf{73.98\%} & \textbf{20.57\%} & \textbf{79.42\%}  & \textbf{15.13\%} & \textbf{76.43\%} & \textbf{18.12\%}
\\
\toprule
\end{tabular}}
\label{tab:bgs}
\end{table*}

\begin{figure*}[ht]
    \centering
    \includegraphics[height=7.5cm]{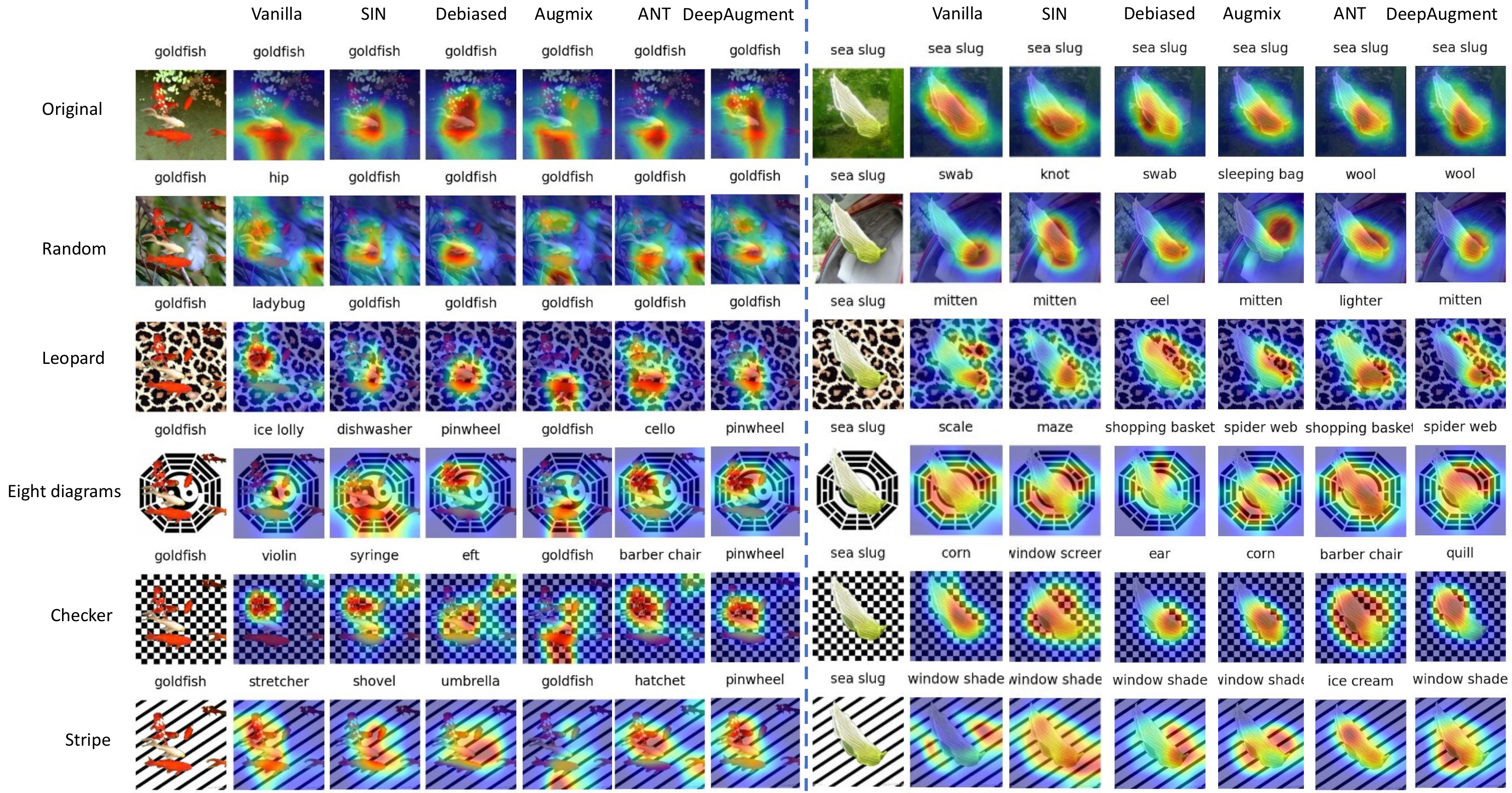}
\caption{Heat maps under different backgrounds.}
    % \vspace{-5mm}
    \label{fig:heat_map_bgs}
\end{figure*}

\section{Further discussion on the distribution}
In this paper, our effort aims to give an editable image tool that can edit the object's attribute in the given image while maintaining it in the original distribution for model debugging.
Thus, we choose the out-of-distribution~(OOD) detection methods including Energy~\cite{liu2020energy} and GradNorm~\cite{huang2021importance} following DRA~\cite{DRA} as the evaluation methods to find out whether our editing tool will move the edited image out of its original distribution.
In contrast to FID which indicates the divergence of two datasets, the OOD detection is used to indicate the extent of the deviance of a single input image from the in-distribution dataset.
% Here we provide a comprehensive evaluation on each kind of attribute editing in Figure~\ref{fig:OOD_each_trans}.
% Here We provide further comparisons with other datasets including ImageNet-V2, inpainted ImageNet-S, adversarial examples and ImageNet-9. With the results in Figure~\ref{fig:OOD_appendix}, we can find that the ImageNet-E holds the best proximity to the ImageNet. This implies that our editing tool can give a controlled evaluation on object attribute changes.}

Covariate shift adaptation(\textit{a.k.a} batch-norm adaptation, BNA) is a way for improving robustness against common corruptions~\cite{schneider2020improving}. Thus, it can help to get a top-1 accuracy performance boost in OOD data. One can easily find out if the provided dataset is OOD by checking whether the BNA can get a performance boost on its data. We have tested the full adaptation results using BNA on ResNet50. In contrast to the promotion on other out-of-distribution dataset, we find that this operation induces little changes to top-1 accuracy on both ImageNet validation set ($0.7615 \rightarrow 0.7613$) and our ImageNet-E ($0.7934 \rightarrow 0.7933$ under $\lambda=20$, $0.6521 \rightarrow 0.6514$ under random position scenario, mean accuracy of 5 runs).  This similar tendency implies that our ImageNet-E shares a similar property with ImageNet.

% \begin{figure}[ht]
%     \centering
%     \includegraphics[height=4.5cm]{iclr2023/Figures/OOD_each_trans.png}
% \caption{\textcolor{blue}{The distribution of the OOD scores for in-distribution (ImageNet) and each transformation. Higher overlap indicates greater proximity to ImageNet.}}
%     % \vspace{-5mm}
%     \label{fig:OOD_each_trans}
% \end{figure}

% \begin{figure*}[ht]
%     \centering
%     \includegraphics[width=\textwidth]{Figures/OOD_appendix_2.pdf}
% \caption{\textcolor{blue}{The distribution of the OOD scores for in-distribution (ImageNet) and other datasets.
% Higher overlap indicates greater proximity to ImageNet.}}
%     % \vspace{-5mm}
%     \label{fig:OOD_appendix}
% \end{figure*}

\section{Further evaluation on more state-of-the-art models}
To provide evaluations on more state-of-the-art models, we step further to evaluate the CLIP~\cite{CLIP} and EfficientNet-L2-Noisy-Student~\cite{NT}. 
The average dropped accuracy in terms of different models can be found in Figure~\ref{fig:top1_DA}.
CLIP shows a good robustness to out-of-distribution data~\cite{kumar2022fine}. Therefore, to find out whether the CLIP can also show a good robustness against attribute editing, we evaluate the CLIP model~(Backbone ViT-B) with both the zero-shot and end-to-end finetuned version. To achieve this, we finetune the pretrained CLIP on the ImageNet training dataset based on prompt-initialized model following \cite{wortsman2022robust}. It acquires a 81.2\% top-1 accuracy on ImageNet validation set while it is 68.3\% for zero-shot version. 
The evaluation on ImageNet-E is shown in Table~\ref{tab:CLIP_OOD} and Table~\ref{tab:CLIP}. 
% It seems that the contrastive learning-based models including MoCo and DINO in Table~\ref{tab:mae} usually show worse robustness to attribute editing.
Though previous studies have shown that the zero-shot CLIP model exhibits better out-of-distribution robustness than the finetuned ones, the finetuned CLIP shows better attribute robustness on ImageNet-E, as shown in Table~\ref{tab:CLIP_OOD} and Table~\ref{tab:CLIP}. The tendency on ImageNet-E is the same with ImageNet~(IN) validation set and ImageNet-V2~(IN-V2). This implies that the ImageNet-E shows a better proximity to ImageNet than other out-of-distribution benchmarks such as ImageNet-C~(IN-C), ImageNet-A~(IN-A). 
Another finding is that the CLIP model fails to show better robustness than ViT-B while they share the same architectures. We suspect that this is caused by that CLIP may have spared some capacity for out-of-distribution robustness. As the network gets larger, its attribute robustness gets better.

While EfficientNet-L2-Noisy-Student is one of the top models on ImageNet-A benchmark~\cite{NT}, it also shows superiority on ImageNet-E. To delve into the reason behind this, we test EfficientNet-L2-Noisy-Student-475~(EF-L2-NT-475) and EfficientNet-B0-Noisy-Student~(EF-B0-NT). The EF-L2-NT-475 differs from EF-L2-NT in terms of input size, which former is 475 while it is 800 for the latter. It can be found that the input size can induce little improvement to the attribute robustness. In contrast, larger networks can benefit a lot to attribute robustness, which is consistent with the finding in Section 5.1. 
% When comparing EF-B0-NT with EF-B0, we find that the noisy student training induces little improvement to attribute robustness.

Evaluations on $91$ state-of-the-art models can be found in Figure~\ref{fig:91}. All the evaluated models in this figure are all provided by the timm library, except for the MoCo-V3-FT and CLIP-FT, which are finetuned by us.
% They are end-to-end finetuned by us since we find that the pretrained models fail to show a good attribute robustness. For example, compared to the officially released MAE, the MoCo-v3 fails to improve the robustness against attribute changes compared to the vanilla ViT-B/16. When we dive into the reason, we find that the officially released MoCo-v3 lacks the procedure of end-to-end finetuning. It only achieves 0.9318 top-1 accuracy on the original images. Therefore, to find out whether the contrastive learning based self-supervised methods can help for the attribute robustness, we finetuned it and the attribute robustness has gained a boost in attribute robustness.

\begin{table*}[ht]
% \color{blue}
\caption{More evaluations on state-of-the-art models including CLIP and EfficientNet-L2-Noisy-Student.}
\resizebox{1.0\linewidth}{!}{
\begin{tabular}{c|c|ccccc|cccc|c|c|c}
\toprule
\multirow{2}{*}{Models} & \multirow{2}{*}{Ori} &  \multicolumn{5}{c|}{Background changes}                                                   & \multicolumn{4}{c|}{Size changes}                                  & Position   & Direction   & \multirow{2}{*}{Avg.}    \\ \cline{3-13}
                               &                      & Inver & $\lambda=-20$            & $\lambda=20$            & $\lambda=20$-Adv           & Random            & Full    & 0.1              & 0.08             & 0.05              & rp   & rd    &         \\ \hline\hline
% ViT-B        & \textbf{0.9566} & \textbf{0.71\%} & \textbf{5.44\%} & \textbf{8.37\%} & \textbf{12.42\%} & \textbf{25.35\%} & 0.52\%    & \textbf{4.91\%} & \textbf{6.87\%} & \textbf{16.4\%} & \textbf{20.35\%} & \textbf{11.96\%} \\
% CLIP-zero-shot & 0.8001 & 6.10\% & 14.44\% & 19.10\% & 24.70\% & 45.17\% & 4.16\% & 15.85\% & 19.72\% & 31.63\% & 36.08\% & 26.96\% \\
% CLIP-FT         & 0.9368          & 2.32\%          & 10.49\%         & 12.63\%         & 18.31\%          & 40.92\%          & 4.65\%    & 9.87\%          & 13.52\%         & 24.89\%         & 30.48\%          & 23.49\%          \\ \hline
% EF-B0        & 0.9285          & 1.15\%          & 7.65\%          & 11.54\%         & 27.08\%          & 37.57\%          & 3.54\%    & 8.61\%          & 12.46\%         & 25.08\%         & 30.06\%          & 20.58\%          \\
% EF-B0-NT     & 0.9430          & 2.08\%          & 8.94\%          & 11.15\%         & 21.03\%          & 37.05\%          & 1.22\%    & 8.39\%          & 12.20\%         & 24.35\%         & 29.29\%          & 20.22\%            \\
% EF-L2-NT-475 & \textbf{0.9784} & \textbf{1.10\%} & 3.68\%          & 4.61\%          & 5.99\%           & 15.21\%          & \textbf{0.52\%}    & \textbf{2.26\%} & \textbf{2.77\%} & 5.62\%          & 7.52\%           & 4.68\%           \\
% EF-L2-NT     & 0.9763          & 1.29\%          & \textbf{3.58\%} & \textbf{4.15\%} & \textbf{5.32\%}  & \textbf{13.03\%} & 0.73\%    & 2.32\%          & 2.85\%          & \textbf{5.13\%} & \textbf{6.17\%}  & \textbf{4.65\%} \\ \toprule
ViT-B/16                   & 95.38\%                                       & \textbf{0.83\%}                            & 5.32\%                            & 8.43\%                           & 26.60\%                              & 10.98\%                    & \textbf{0.62\%}          & 4.00\%                  & 6.30\%                   & 14.51\%                  & 18.82\%                      & 14.95\%                       & 11.05\%               \\ \hline
\multicolumn{14}{c}{Zero-shot}                                                                                                                                \\ \hline
CLIP\_RN50              & 72.38\%                                       & 6.03\%                                     & 11.64\%                           & 16.72\%                          & 35.07\%                              & 21.82\%                    & 8.78\%                   & 14.39\%                 & 17.69\%                  & 26.48\%                  & 29.79\%                      & 25.31\%                       & 20.77\%               \\
CLIP\_RN101             & 73.35\%                                       & 4.51\%                                     & 10.77\%                           & 14.42\%                          & 33.42\%                              & 19.63\%                    & 6.39\%                   & 14.53\%                 & 18.19\%                  & 26.58\%                  & 30.08\%                      & 24.51\%                       & 19.85\%               \\
CLIP\_RN50x4            & 77.18\%                                       & 4.64\%                                     & 10.44\%                           & 13.27\%                          & 31.39\%                              & 18.51\%                    & 7.46\%                   & 12.37\%                 & 15.66\%                  & 24.23\%                  & 27.19\%                      & 24.25\%                       & 18.48\%               \\
CLIP\_RN50x16           & 82.10\%                                       & 4.39\%                                     & 10.10\%                           & 12.41\%                          & 27.14\%                              & 16.62\%                    & 6.62\%                   & 11.10\%                 & 13.53\%                  & 22.09\%                  & 25.27\%                      & 23.13\%                       & 16.80\%               \\
CLIP\_RN50x64           & 85.66\%                                       & 4.77\%                                     & \textbf{8.89\%}                   & \textbf{10.79\%}                 & \textbf{23.75\%}                     & \textbf{13.44\%}           & 6.39\%                   & 9.20\%                  & 11.92\%                  & \textbf{19.17\%}         & \textbf{21.62\%}             & 20.57\%                       & \textbf{14.57\%}      \\
CLIP\_ViT-B/32          & 74.08\%                                       & 5.55\%                                     & 13.24\%                           & 18.64\%                          & 43.26\%                              & 26.39\%                    & 2.99\%                   & 15.59\%                 & 19.74\%                  & 29.05\%                  & 33.37\%                      & 24.89\%                       & 22.72\%               \\
CLIP\_ViT-B/16          & 80.01\%                                       & 4.88\%                                     & 11.56\%                           & 15.28\%                          & 36.14\%                              & 20.09\%                    & 4.88\%                   & 12.67\%                 & 15.77\%                  & 25.31\%                  & 28.87\%                      & 21.57\%                       & 19.21\%               \\
CLIP\_ViT-L/14          & 87.61\%                                       & 4.35\%                                     & 11.04\%                           & 14.46\%                          & 33.69\%                              & 18.35\%                    & \textbf{1.81\%}          & 11.67\%                 & 15.09\%                  & 23.66\%                  & 27.19\%                      & 18.05\%                       & 17.50\%               \\
CLIP\_ViT-L/14-336      & \textbf{88.01\%}                              & \textbf{3.16\%}                            & 9.07\%                            & 12.25\%                          & 29.69\%                              & 16.08\%                    & 3.16\%                   & \textbf{9.20\%}         & \textbf{11.78\%}         & 19.94\%                  & 22.89\%                      & \textbf{16.15\%}              & 15.02\%               \\
CLIP\_ViT-L/14-336      & \textbf{88.01\%}                              & \textbf{3.16\%}                            & 9.07\%                            & 12.25\%                          & 29.69\%                              & 16.08\%                    & 3.16\%                   & \textbf{9.20\%}         & \textbf{11.78\%}         & 19.94\%                  & 22.89\%                      & \textbf{16.15\%}              & 15.02\%               \\
\hline
\multicolumn{14}{c}{Finetune}                                                                                                                                                                                                                                                                                                                                                                                                                                     \\
\hline
CLIP\_ViT-B/16-FT       & 93.68\%                                       & 2.17\%                                     & 9.82\%                            & 11.83\%                          & 38.33\%                              & 18.19\%                    & 4.66\%                   & 9.25\%                  & 12.67\%                  & 23.32\%                  & 28.56\%                      & 22.00\%                       & 17.86\%               \\
CLIP\_ViT-L/14-336-FT   & \textbf{96.97\%}                              & 1.29\%                                     & \textbf{5.16\%}                   & \textbf{6.18\%}                  & \textbf{19.93\%}                     & \textbf{8.09\%}            & 1.29\%                   & \textbf{3.47\%}         & \textbf{4.90\%}          & \textbf{10.98\%}         & \textbf{13.74\%}             & \textbf{10.96\%}              & \textbf{8.47\%}       \\
                        \hline
EF-B0                   & 92.85\%                                       & \textbf{1.07\%}                            & 7.10\%                            & 10.71\%                          & 34.88\%                              & 15.64\%                    & 3.03\%                   & 8.00\%                  & 11.57\%                  & 23.28\%                  & 27.91\%                      & 19.11\%                       & 16.12\%               \\
EF-B0-NT                & 94.30\%                                       & 1.97\%                                     & 8.43\%                            & 10.51\%                          & 34.93\%                              & 15.99\%                    & 1.79\%                   & 7.91\%                  & 11.50\%                  & 22.96\%                  & 27.62\%                      & 19.07\%                       & 16.07\%               \\
EF-B7                   & 97.10\%                                       & 1.80\%                                     & 6.37\%                            & 7.20\%                           & 23.36\%                              & 10.78\%                    & 1.65\%                   & 4.16\%                  & 6.25\%                   & 14.13\%                  & 17.12\%                      & 10.56\%                       & 10.16\%               \\
EF-B7-NT                & 97.38\%                                       & 1.30\%                                     & 5.26\%                            & 6.10\%                           & 19.96\%                              & 9.15\%                     & 0.55\%                   & 3.31\%                  & 4.75\%                   & 10.67\%                  & 12.87\%                      & 7.98\%                        & 8.06\%                \\
EF-L2-NT-475            & \textbf{97.84\%}                              & 1.08\%                                     & 3.60\%                            & 4.51\%                           & 14.88\%                              & 7.14\%                     & 0.51\%                   & 2.21\%                  & \textbf{2.71\%}          & 5.50\%                   & 7.35\%                       & 4.58\%                        & 5.30\%                \\
EF-L2-NT                & 97.63\%                                       & 1.26\%                                     & \textbf{3.50\%}                   & \textbf{4.06\%}                  & \textbf{12.73\%}                     & \textbf{6.90\%}            & \textbf{0.71\%}          & \textbf{2.27\%}         & 2.79\%                   & \textbf{5.01\%}          & \textbf{6.03\%}              & \textbf{4.55\%}               & \textbf{4.85\%}      
\\ \toprule
\end{tabular}}
\label{tab:CLIP}
\end{table*}

\section{Failure cases of generated images}
\begin{figure*}[t]
    \centering
    \includegraphics[width=\textwidth]{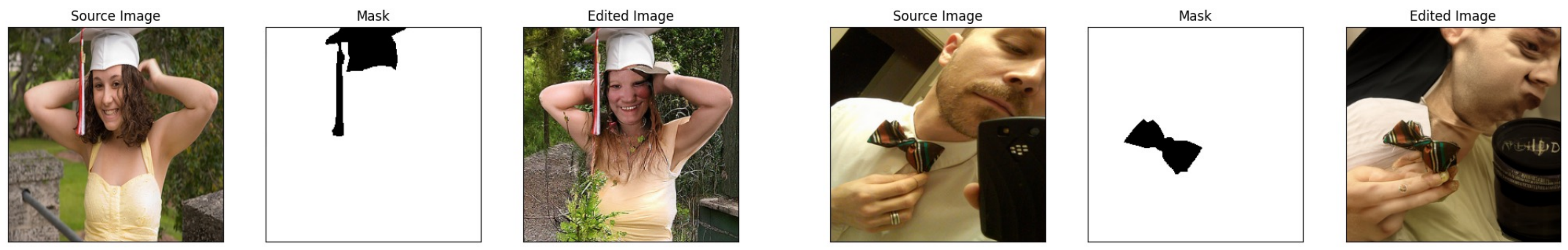}
\caption{The failure cases of attribute editing.}
    % \vspace{-5mm}
    \label{fig:Failure_of_generation}
\end{figure*}

The failure cases of generated images are shown in Figure~\ref{fig:Failure_of_generation}. The diffusion model fails to generate high-quality person images. 
Though the object is reserved, the whole image looks quite wired.
Therefore, we only keep the animal classes, resulting a compact set of ImageNet-E. 
However, extensive evaluations to $919$ in Section~\ref{sec:A_more_data} have witnessed a same conclusion with evaluations on 373 classes. This implies that using our ImageNet-E can already reflect the model robustness against object attribute changes.

\section{Related literature to robustness enhancements}
\textbf{Adversarial training}. 
\cite{salman2020adversarially} focus on adversarially robust ImageNet classifiers and show that they yield improved accuracy on a standard suite of downstream classification tasks. It provides a strong baseline for adversarial training. Therefore, we choose their officially released adversarially trained models\footnote{https://github.com/microsoft/robust-models-transfer} as the evaluation model. Models with different architectures are adopted here\footnote{https://github.com/alibaba/easyrobust}.

\textbf{SIN}~\cite{SIN} 
% shows that ImageNet-trained CNNs are strongly biased towards recognizing textures rather than shapes and proposes to train the model on `Stylized-ImageNet' to learn a shape-based representation. 
provides evidence that machine recognition today overly relies on object textures rather than global object shapes, as commonly assumed. It demonstrates the advantages of a shape-based representation for robust inference (using their Stylized-ImageNet dataset to induce such a representation in neural networks) 

\textbf{Debiased}~\cite{debiased} shows that convolutional neural networks are often biased towards either texture or shape, depending on the training dataset, and such bias degenerates model performance. Motivated by this observation, it develops a simple algorithm for shape-texture Debiased learning. To prevent models from exclusively attending to a single cue in representation learning, it augments training data with images with conflicting shape and texture information (\textit{e.g.}, an image of chimpanzee shape but with lemon texture) and provides the corresponding supervision from shape and texture simultaneously. It empirically demonstrates the advantages of the shape-texture Debiased neural network training on boosting both accuracy and robustness.

\textbf{Augmix}~\cite{hendrycks2020Augmix}
focuses on the robustness improvement to unforeseen data shifts encountered during deployment. It proposes a data processing technique named Augmix that helps to improve robustness and uncertainty measures on challenging image classification benchmarks.

\textbf{ANT}~\cite{ANT}
demonstrates that a simple but properly tuned training with additive Gaussian and Speckle noise generalizes surprisingly well to unseen corruptions, easily reaching the previous state of the art on the corruption benchmark ImageNet-C and on MNIST-C.

\textbf{DeepAugment}~\cite{DeepAugment}.
Motivated by the observation that using larger models and artificial data augmentations can improve robustness on real-world distribution shifts, contrary to claims in prior work. It introduces a new data augmentation method named DeepAugment, which uses image-to-image neural networks for data augmentation. It improves robustness on their newly introduced ImageNet-R benchmark and can also be combined with other augmentation methods to outperform a model pretrained on 1000× more labeled data.

\bigskip
There are some more tables and figures in the next pages.

\end{document}